\documentclass{article}

\usepackage{microtype}
\usepackage{graphicx}
\usepackage{amsfonts, amsmath}
\usepackage[hidelinks]{hyperref}
\usepackage{subfigure}
\usepackage[textsize=tiny]{todonotes}
\usepackage{booktabs} 

\usepackage{xcolor}
\usepackage{float}        
\usepackage{caption}      
\usepackage[most]{tcolorbox}    
\usepackage{hyperref}     
\usepackage{enumitem}
\usepackage{pifont}
\usepackage{tablefootnote}
\usepackage{array}
\usepackage{multirow}

\newfloat{prompt}{htbp}{lop}[section] 
\floatname{prompt}{Prompt}            

\tcbset{
    promptbox/.style={
        colback=gray!10,      
        colframe=gray!50,     
        boxrule=0.5pt,        
        arc=4pt,              
        left=6pt,             
        right=6pt,            
        top=6pt,              
        bottom=6pt,           
        enhanced,             
    }
}

\usepackage[accepted]{icml2025}

\begin{document}

\twocolumn[
\icmltitle{
Al-Khwarizmi: Discovering Physical Laws with Foundation Models
}

\begin{icmlauthorlist}
\icmlauthor{Christopher E. Mower}{noah}
\icmlauthor{Haitham Bou-Ammar}{noah,ucl}
\end{icmlauthorlist}

\icmlaffiliation{noah}{Huawei's Noahs Ark Lab, London, UK.}
\icmlaffiliation{ucl}{University College London, London, UK}

\icmlcorrespondingauthor{Christopher E. Mower}{christopher.mower@huawei.com}

\vskip 0.3in
]

\printAffiliationsAndNotice{} 

\begin{abstract}
Inferring physical laws from data is a central challenge in science and engineering, including but not limited to healthcare, physical sciences, biosciences, social sciences, sustainability, climate, and robotics. 
Deep networks offer high-accuracy results but lack interpretability, prompting interest in models built from simple components. 
The Sparse Identification of Nonlinear Dynamics (SINDy) method has become the go-to approach for building such modular and interpretable models. 
SINDy leverages sparse regression with L1 regularization to identify key terms from a library of candidate functions. 
However, SINDy's choice of candidate library and optimization method requires significant technical expertise, limiting its widespread applicability.
This work introduces Al-Khwarizmi, a novel agentic framework for physical law discovery from data, which integrates foundational models with SINDy. 
Leveraging LLMs, VLMs, and Retrieval-Augmented Generation (RAG), our approach automates physical law discovery, incorporating prior knowledge and iteratively refining candidate solutions via reflection. 
Al-Khwarizmi operates in two steps: it summarizes system observations—comprising textual descriptions, raw data, and plots—followed by a secondary step that generates candidate feature libraries and optimizer configurations to identify hidden physics laws correctly. 
Evaluating our algorithm on over \emph{198 models}, we demonstrate \emph{state-of-the-art performance} compared to alternatives, reaching a \emph{$20\%$ increase against the best-performing alternative.}
\end{abstract}

\section{Introduction}

A central challenge in science and engineering is inferring physical laws, or dynamical systems, from data. 
This includes important problems like understanding protein folding, predicting earthquake aftershocks, optimizing traffic flow, enhancing crop yields, and improving renewable energy efficiency. 
Recent advances in artificial intelligence~\cite{krizhevsky2012imagenet, LeCun2015, Mnih2015, Vaswani17} have fueled enthusiasm for its application in accelerating scientific discovery across fields such as material science~\cite{Fang18}, biology~\cite{Jumper2021}, mathematics~\cite{Davies2021}, nuclear fusion~\cite{Degrave2022}, and data science~\cite{grosnit2024largelanguagemodelsorchestrating}.

Many phenomena can be described by $\dot{x} = f(x)$, where $x$ is the state, $\dot{x}$ its time derivative, and $f$ is a potentially nonlinear function. 
The goal of \textit{physical law discovery} is to approximate $f$ with $\widehat{f}$. 
Historically, $\widehat{f}$ was derived from first principles, but as real-world measurements became more accessible, \textit{data-driven methods} emerged, using regression to fit a model $\widehat{f}_\theta$ by minimizing a loss function. 
Due to limited computational resources, early methods often relied on linear models $\widehat{f}_\theta(x) = \theta x$.

\begin{figure}
    \centering
    \includegraphics[width=\linewidth]{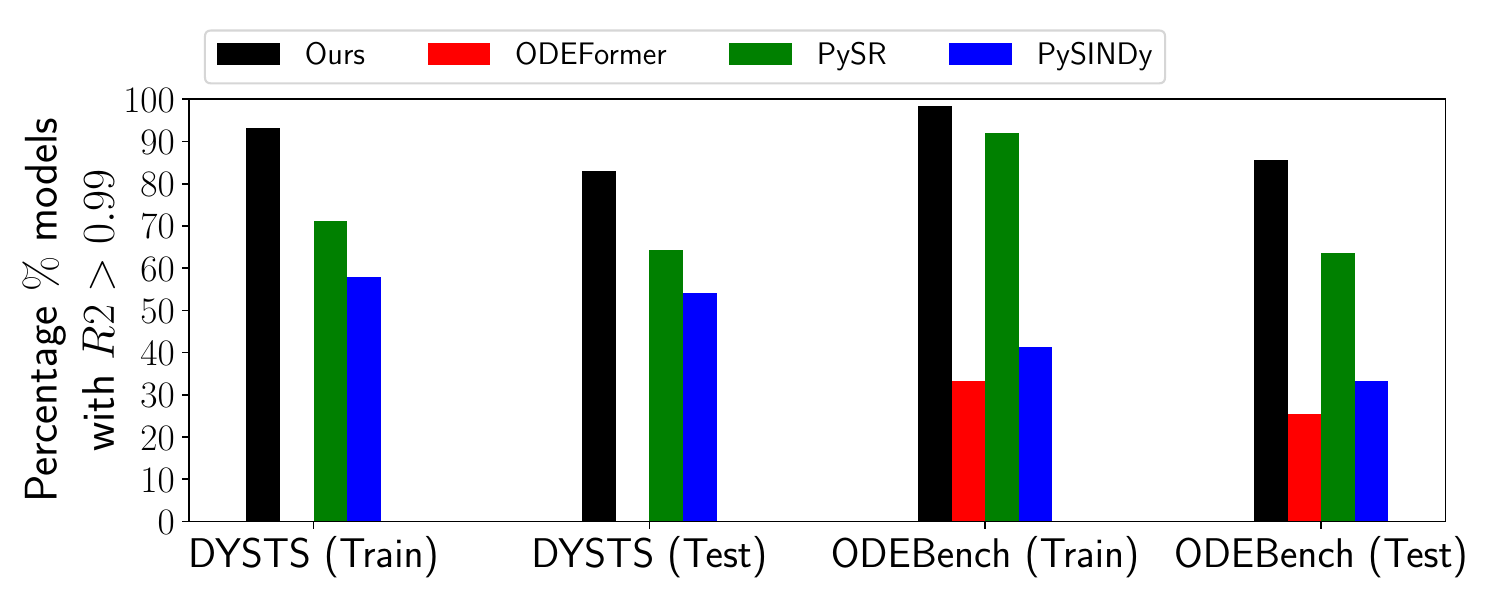}
    \caption{Comparison of our approach with alternatives on DYSTS and ODEBench, showing the percentage of models with~$R2 > 0.99$ on training and test data.}
    \label{fig:overal}
\end{figure}

With 
(i) the exponential growth of computing power~\cite{Nordhaus_2007, Thompson23}, 
(ii) greater hardware accessibility through languages like FORTRAN, C, C++, and Python~\cite{huang2024programming}, and 
(iii) the development of specialized open-source libraries such as LAPACK, NumPy, and PyTorch~\cite{Langenkamp22}, 
modern methods, especially those based on deep learning~\cite{LeCun2015}, have become highly effective~\cite{wehmeyer2018time, mardt2018vampnets, vlachas2018data, pathak2018model, raissi2019physics, champion2019data, raissi2020hidden, yang2020physics, lu2021deepxde}.

\begin{figure*}[t]
    \centering
    \includegraphics[width=\linewidth]{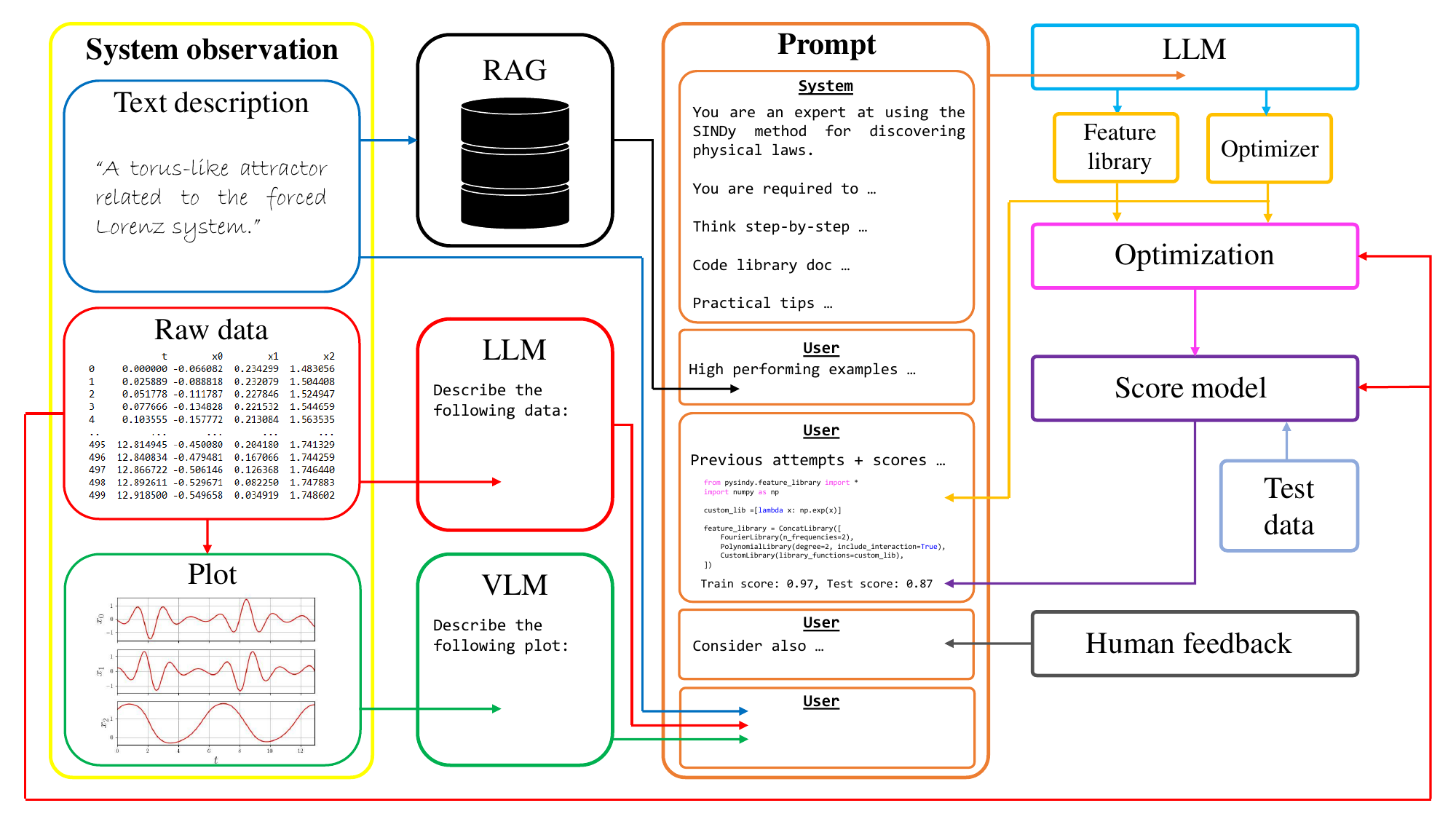}
    \caption{An overview of our proposed approach for governing dynamics discovery with foundation models.}
    \label{fig:method-overview}
\end{figure*}

While deep learning excels in forecasting complex nonlinear systems, its lack of interpretability~\cite{Chakraborty17} presents a challenge. 
Recent efforts have focused on developing \textit{parsimonious} models~\cite{Bongard07, Schmidt09, Brunton2016} that balance accuracy, simplicity, and interpretability. 
A historical example of this is Kepler’s data-driven model of planetary motion, which, while accurate, lacked generalizability, unlike Newton’s laws, which provided a deeper, universal understanding and enabled advancements like the 1969 moon landing. 
This underscores the need for models that not only predict effectively but also offer a deeper physical understanding. 
Additionally, deep learning's reliance on large datasets makes methods like Sparse Identification of Nonlinear Dynamics (SINDy)~\cite{Brunton16b, kaiser2018sparse}, shown to be effective in low-data scenarios, especially valuable.
Moreover, deep learning demands specialized expertise, a rapidly growing economic need~\cite{ALEKSEEVA2021102002}.

Recent breakthroughs~\cite{Vaswani17} have led to the development of foundation models such as 
LLaMA~\cite{touvron2023}, 
GPT-4~\cite{openai2024gpt4technicalreport}, 
Qwen~\cite{bai2023}, 
Falcon~\cite{almazrouei2023}, 
LLaVA~\cite{liu2023improvedllava, liu2023llava}, and 
InternVL~\cite{chen2024}. 
These models have demonstrated remarkable generalization and reasoning capabilities, and thus spurred applications across fields including 
medicine~\cite{Moor2023}, 
education~\cite{Tianlong24}, 
law~\cite{henderson2023foundation}, and 
robotics~\cite{brohan2023can, mower2024ros}. 
Recently,~\citet{Mengge24} proposed LLM4ED, which uses LLMs for physical law discovery in a symbolic regression setup with reflection to suggest candidate models $\widehat{f}_\theta$. 
While impressive, this method does not incorporate prior knowledge that could enhance convergence and overlooks critical factors like the choice and configuration of the optimizer.

We introduce Al-Khwarizmi, a modular framework for data-driven physical law discovery that combines foundation models with the SINDy method, as shown in Figure~\ref{fig:method-overview}. 
While SINDy effectively builds interpretable models through sparse regression with L1 regularization, its reliance on expert knowledge limits accessibility. 
In contrast, our framework utilizes foundation models like LLMs and VLMs to provide the expert knowledge. 
This is especially valuable given the growing demand for deep learning expertise~\cite{ALEKSEEVA2021102002}. 
Al-Khwarizmi also integrates prior knowledge via Retrieval-Augmented Generation (RAG)~\cite{NEURIPS2020_6b493230} and refines solutions through reflection, inspired by \cite{Mengge24}. 
Extensive experiments show that our method outperforms existing approaches on two benchmarks, as summarized in Figure~\ref{fig:overal}.
Moreover, all the results from our framework use \textit{only} open-source models.

Pretrained foundation models possess specialized capabilities based on their training data. 
For example, VLMs are trained on image captions, while LLMs focus on text data like novels, spreadsheets, and code. 
Consequently, these models store distinct commonsense knowledge. 
Al-Khwarizmi leverages this by operating in two steps: 
first, system observations—text descriptions, raw data, and plots—are summarized using LLMs and VLMs. 
Then, candidate feature libraries and optimizer configurations are generated, a SINDy model is optimized under the candidate sample, and compared with measured data, iterating until convergence.

Al-Khwarizmi draws inspiration from expert data interpretation, often through plots to identify patterns. 
Similarly, it uses a VLM to extract insights from plots of measured data, leveraging the commonsense knowledge embedded in foundation models. 
The potential of foundation models for logical reasoning with chart data remains under-explored~\cite{xia2024chartx}, and our work advances this area. 
While raw data, especially at high frequencies, can exceed a model's context length, we find summarizing the data with an LLM in Al-Khwarizmi’s initial step is highly effective. 
LLMs are trained on datasets including spreadsheets and thus our work supports the hypothesis that they are equipped to reason about such data.
Moreover, expert-written descriptions, practical advice, and code documentation provide valuable prior knowledge, guiding the foundation model through reflection iterations that refine model selection and improve the discovery of physical laws. 
We also provide evidence that feedback from non-experts can further enhance the framework's performance.

The following is a summary of our main contributions.
\begin{itemize}
    \item We propose a modular agentic framework for physical law discovery utilizing pretrained foundation models to analyze system observations and guide the the SINDy method.
    \item Our work shows that RAG, reflection, and human feedback (expert and non-expert) can be incorporated to further improve solutions without fine-tuning the foundation model.
    \item We conduct extensive experiments on two benchmarks containing a combination of $63+135=198$ models.
    \item Our code, datasets, prompts, and results can be made available upon request.
\end{itemize}

\section{Problem formulation}\label{eq:problem-formulation}

A dynamical system modeling a physical law is typically represented as
\begin{equation}\label{eq:governing-dynamics}
    \dot{x} = f(x)\quad\text{with}\quad f:\mathbb{R}^n\rightarrow\mathbb{R}^n
\end{equation}
where 
$x \in \mathbb{R}^n$ is the $n$-dimensional system state, 
$\dot{x} \equiv \partial x/\partial t$ is the elementwise time derivative of $x$, and 
$f$ represents the governing dynamics (or \textit{right-hand side function}). 
In \eqref{eq:governing-dynamics}, $f$ is unknown, and the goal is to identify a suitable model that accurately approximates it from data.

The data used to approximate $f$ we refer to as the \textit{system observation} and denote by $\mathcal{O}$. We frame physical law discovery as a maximum likelihood problem:
\begin{equation}
    \label{eq:governing-dynamics-discovery}
    \widehat{f} = \underset{f_c\in\mathbb{F}}{\text{arg}\max}~p(f_c~|~\mathcal{O})~~\Leftrightarrow~~\widehat{f}\leftarrow A_\psi(\mathcal{O}) 
\end{equation}
where $f_c \in \mathbb{F}$ is a candidate function from some function space $\mathbb{F}$, and $p(\cdot | \cdot)$ is the likelihood that $f_c$ models $f$ given $\mathcal{O}$. The right-hand side shows an equivalent formulation as a mapping, with $A_\psi$ representing an algorithm, e.g., \textsc{Adam}~\cite{kingma2017adam}, and $\psi \in \mathbb{R}^h$ as a set of $h$ hyperparameters (e.g., step size). Choosing $A_\psi$ and $\psi$ often requires expert knowledge.

In previous literature, $\mathcal{O}$ is often a set of $N$ measurements
\begin{equation}\label{eq:system-observation-raw-measurements}
\mathcal{D} := \Big\{\big[(\widetilde{t}_0, \widetilde{x}_0), \dots, (\widetilde{t}_{K_i}, \widetilde{x}_{K_i})\big]\Big\}_{i=1}^N
\end{equation}
where 
$K_i$ is the number of points in the $i$'th trajectory, that may vary. 
Given some parametrization $f_\theta$ (e.g., neural network), we can reformulate~\eqref{eq:governing-dynamics-discovery} as a numerical optimization program, given by
\begin{equation}
    \label{eq:governing-dynamics-discovery-regression}
    \widehat{\theta} = \underset{\theta}{\text{arg}\min}~\mathcal{L}(\theta; \mathcal{O})
\end{equation}
where $\mathcal{L}(\cdot)$ is a loss function (e.g., mean squared error).
Thus, methods based on line-search or trust-regions can then be employed to find $\widehat{\theta}$.

Visualizing data as diagrams is more intuitive for humans and often essential for informed modeling decisions. 
Libraries such as NumPy and Matplotlib facilitate plotting data $\mathcal{D}$. 
Pretrained foundation models can process diverse data types, enabling integration of raw data, text, and images—where text provides context and images reveal insights beyond raw measurements.


In our case, we assume $\mathcal{D}$ can be mapped to images $\mathcal{I}$ (i.e., plots) and also additional textual information $\mathcal{T}$ are available.
The system observation is thus assumed to be given by 
\begin{equation}\label{eq:system-observation}
    \mathcal{O} := (\mathcal{T}, \mathcal{D}, \mathcal{I}).
\end{equation}

Our goal in this work is to demonstrate that AI systems can be utilized to infer $\widehat{f}$ by 
i) utilizing multiple data modalities (i.e., raw data, text, and images) and
ii) providing automated choices of $A_\psi$ and $\psi$ to solve \eqref{eq:governing-dynamics-discovery}.

\section{Related work}

This section reviews relevant works addressing the problem outlined in section \ref{eq:problem-formulation}, including common data-driven methods and recent transformer-based approaches.

\subsection{Data-driven techniques}

Discovering dynamical system models from data is crucial in fields such as science and engineering.
While traditional models are derived from first principles, this approach can be highly challenging in areas like climate science, finance, and biology. 
Data-driven methods for physical law discovery are an evolving field, 
with various techniques including
linear methods \cite{nellesnonlinear, ljung2010perspectives}, 
dynamic mode decomposition (DMD)~\cite{schmid2010dynamic, kutz2016dynamic}, 
nonlinear autoregressive models~\cite{akaike1969autoregressive,billings2013nonlinear}, 
neural networks~\cite{yang2020physics,wehmeyer2018time,mardt2018vampnets,vlachas2018data,pathak2018model,lu2021deepxde,raissi2019physics,champion2019data,raissi2020hidden}, 
Koopman theory~\cite{budivsic2012applied,mezic2013analysis,williams2015data,klus2018data},
nonlinear Laplacian spectral analysis~\cite{giannakis2012nonlinear}, 
Gaussian process regression~\cite{raissi2017machine,raissi2018hidden}, 
diffusion maps~\cite{yair2017reconstruction}, 
genetic programming~\cite{daniels2015automated,schmidt2009distilling,bongard2007automated}, and 
sparse regression~\cite{Brunton2016,rudy2017data,schaeffer2017learning}, among other recent advances.
Sparse regression techniques, such as SINDy~\cite{Brunton2016}, have proven to be an effective method, offering high computational efficiency and a straightforward methodology. 
SINDy has demonstrated strong performance across various fields~\cite{Shea_2021,MESSENGER2021110525,kaheman2020sindy,Fasel_2022}. 
However, in order to implement effectively, it relies on prior knowledge and technical expertise. 
In this work, we leverage foundation models to supply this missing prior knowledge and expertise.

\subsection{Transformer-based discovery}

The introduction of transformer models~\cite{Vaswani17} has made it possible to learn sequence-to-sequence tasks in a broad range of domains. 
Combined with large-scale pre-training, transformers have been successfully applied to symbolic tasks, including 
function integration~\cite{Lample2020Deep}, 
logic~\cite{hahn2021teaching}, and theorem proving~\cite{polu2020}.

Recent studies have applied transformers to symbolic regression~\cite{kamienny2022endtoend, Landajuela22, Vastl24}, where transformers are used to predict function structures from measurement data. 
The latest contribution in this line of work is ODEFormer~\cite{ascoli2024odeformer}, a transformer-based model designed to infer multidimensional ordinary differential equations in symbolic form from a single trajectory of observational data. 
In our work, we include ODEFormer in our benchmark comparisons and show our framework is able to find $60\%$ more models with an $R2$ score of above 0.99 on two benchmarks.

In \cite{Mengge24}, an LLM is used for symbolic regression to sample candidate functions $f_c$, optimize their parameters, and score them based on test data. 
This score refines the candidate functions through reflection on prior samples. 
Similar to our approach, \cite{Mengge24} also employs language models in equation discovery, but we enhance it by incorporating system observations like plots and descriptions. 
Additionally, we generate representations not only for the candidate model $\widehat{f}$ but also for the optimizer $A_\psi$ that tunes model parameters. 
Our method is evaluated on a wider range of benchmark problems, showing robustness across diverse data-driven dynamics tasks. 
Importantly, all experiments rely solely on pre-trained, open-source models, without using commercial models like \textsc{GPT-4o}.

\section{Sparse Identification of Nonlinear Dynamical Systems}

SINDy is a data-driven method for discovering physical laws from data~\cite{Brunton2016}.
SINDy constructs~\eqref{eq:system-observation-raw-measurements} as
\begin{subequations}
    \begin{align}
        \widetilde{X} &= \big[ \widetilde{x}_0 | \dots | \widetilde{x}_K \big]^T \in \mathbb{R}^{K \times n},~\text{and}\label{eq:sindy-X}\\
        \dot{\widetilde{X}} &= \big[ \dot{\widetilde{x}}_0 | \dots | \dot{\widetilde{x}}_K \big]^T \in \mathbb{R}^{K \times n}\label{eq:sindy-dX}
    \end{align}
\end{subequations}
where \eqref{eq:sindy-dX} is typically found using finite differencing.
Next, a feature library $\Theta(\cdot)\in\mathbb{R}^{K\times p}$ is constructed including, but not limited to, constant, polynomial, and trigonometric terms
\begin{equation}\label{eq:feature-library}
    \Theta(X) = \Big[1~|~X~|~X^2~|~\cdots~|~\sin(X)~|~\sin(2X)~|~\cdots\Big]
\end{equation}
where $p$ is the total number of primitive functions.
Let $\widehat{\Xi}\in\mathbb{R}^{p\times n}$ be a solution of $\dot{\widetilde{X}} = \Theta(\widetilde{X})\Xi$, found by solving 
\begin{equation}\label{eq:sindy-regression}
    \widehat{\Xi} = \underset{\Xi\in\mathbb{R}^{p\times n}}{\text{arg}\min}~\big\|\Theta(\widetilde{X})\Xi - \dot{\widetilde{X}}\big\|_2 + \lambda \|\Xi\|_1
\end{equation}
where $\|\cdot\|_1, \|\cdot\|_2$ are the 1- and 2-norm respectively,
and $0 < \lambda\in\mathbb{R}$ weights the sparsity objective.
Given $\widehat{\Xi}$, SINDy approximates~\eqref{eq:governing-dynamics} as $\widehat{f}(x) = \widehat{\Xi}^T\Theta(x^T)^T$. 
Extensions of SINDy address multiple trajectories, noisy data, and partial differential equations.
For details on the method and extensions, see~\cite{Brunton16b}, the video series~\cite{brunton2021sindy}, and references therein.

While SINDy is powerful, effective use requires expertise. 
Based on our observations and PySINDy documentation~\cite{deSilva2020}, two key factors are crucial:
the choice of the
(i) feature library~\eqref{eq:feature-library}, and (ii) optimizer for solving~\eqref{eq:sindy-regression}.

\section{Proposed approach}

Now, we present our proposed approach, shown in Figure~\ref{fig:method-overview}. 

\subsection{System observation summarization}\label{sec:summarization}

The system observation~\eqref{eq:system-observation} consists of $\mathcal{T},\mathcal{D},\mathcal{I}$.
Prior work predominantly relied on $\mathcal{D}$.
Instead, we discovered incorporating $\mathcal{T}$ and $\mathcal{I}$ improves performance. 
However, including in the main LLM prompt often exceeded context length, especially when $\mathcal{D}$ is collected at high frequency.


To address this, we introduced a pre-processing step to extract insights from $\mathcal{D}$ and $\mathcal{I}$. 
The raw data is converted to CSV format and summarized by an LLM. 
A JPEG plot, generated using Matplotlib, is summarized by a VLM; e.g, see the Aizawa model~\cite{aizawa1982} in the ``Plot'' part of Figure~\ref{fig:method-overview}. 
Template prompts are given in the Appendix.

\subsection{Retrieval-Augmented Generation}\label{sec:rag}

Methods based on RAG have been shown to improve language model performance~\cite{NEURIPS2020_6b493230}, especially with large knowledge bases~\cite{gao2024}. 
When an example database is available, our framework employs the following RAG-based approach.

We set up a pre-trained text embedding network $e: \mathbb{T} \rightarrow \mathbb{R}^m$ mapping text $\tau \in \mathbb{T}$ to an embedding vector $z \in \mathbb{R}^m$; 
contextually similar text is mapped to nearby regions.
Examples include 
NV-Embed~\cite{lee2024nv}, gte-Qwen~\cite{li2023towards}, and Nomic Embed~\cite{nussbaum2024nomicembedtrainingreproducible}. 

Assuming a database with $R$ example pairs $(d_i, c_i)$: $d_i \in \mathbb{T}$ describes a dynamical system (e.g., in text), and $c_i \in \mathbb{T}$ is the corresponding code defining the feature library and/or optimizer. 
Likeness between descriptions is modeled using the cosine similarity, denoted $s(d_1, d_2) = \frac{e(d_1) \cdot e(d_2)}{\|e(d_1)\| \|e(d_2)\|}$.

Given a query description $q \in \mathbb{T}$ from the system observation~\eqref{eq:system-observation}, we retrieve the top $N$ most similar pairs $\mathcal{J}^*= \{j_1^*, \dots, j_N^*\}$ by maximizing $\sum_{j \in \mathcal{J}} s(q, d_j)$.
The retrieved pairs $\{(d_j, c_j)\}_{j \in \mathcal{J}^*}$ are used to provide additional context and enhance the LLM.

\subsection{Prompt generation}\label{sec:prompt-generation}

The main prompt provided to the LLM for generating the feature library and/or optimizer samples consists of several key components, shown in the center of Figure~\ref{fig:method-overview}. 
Our modular system enables different ablations of these components; summarized below.

\paragraph{Main context}

The main context sets the primary goal for the LLM, guiding its interpretation of user messages and positioning it as an expert in the SINDy method. 
The prompt defines the objective: assisting in generating the feature library and selecting an optimizer within the PySINDy framework. 
It uses chain-of-thought reasoning~\cite{wei2023chainofthoughtpromptingelicitsreasoning} to structure the model’s approach. 
Relevant PySINDy documentation~\cite{deSilva2020, Kaptanoglu2022}, including key classes, parameters, and example usage, is provided. 
Also, we found including practical tips~\footnote{\href{https://pysindy.readthedocs.io/en/latest/tips.html}{https://pysindy.readthedocs.io/en/latest/tips.html}}, enhanced the LLM's performance.

\paragraph{RAG examples}

Given $\mathcal{T}$, $N$ examples are retrieved from the RAG database (see Section~\ref{sec:rag}).
These examples are concatenated and included in the main prompt for the LLM to analyze and help guide its decisions.

\paragraph{Previous attempts}

After each sample is generated from the LLM, a SINDy model is trained, and scores are computed for both training and held-out test data (see Section~\ref{sec:scoring}). 
The best samples, along with their scores, are incorporated into the prompt to aid the model identify factors that lead to improvements. 

\paragraph{Human feedback}

At each iteration, the user can inspect the generated code and plots showing the fitting between the trained SINDy model and training/test data. 
The user is given the chance to provide feedback that is then incorporated into the prompt.

\paragraph{System description}

The system description corresponds to~\eqref{eq:system-observation} in textual form. 
Since the raw data and images were summarized by the respective LLM and VLM (Section~\ref{sec:summarization}), the system description can incorporate ablations of these summaries.
Later, we investigate resulting performance with and without the system description to asses its impact.

\subsection{Model optimization}

Given the main prompt, an LLM produces code for constructing the feature library and/or optimizer; see the Appendix for examples.
Code is extracted from the prompt using regular expressions and saved to disk. 
The feature librar and optimizer are then imported used to train a SINDy model that subsequently solves~\eqref{eq:sindy-regression}.

\subsection{Scoring a model}\label{sec:scoring}

To assess the effectiveness of a trained SINDy model, we use the $R2$ score (Coefficient of Determination); that has also been used in related work~\cite{deSilva2020, Mengge24, ascoli2024odeformer}. 
The $R2$ score is defined by
$R2 = 1 - \sum_{i} (y_i - \widehat{y}_i)^2/\sum_{i} (y_i - \bar{y})^2$ where $y_i$ are the observed values, $\widehat{y}_i$ the predicted values, and $\bar{y}$ the mean of the actual values. 
When $R2=1$ this indicates a perfect fit, $R2=0$ implies no explanatory power, and $R2<0$ suggest worse performance than predicting the mean.


\section{Experiments}

This section reports our experimental findings; further details are found in the Appendix.

\subsection{Preliminaries}\label{sec:preliminaries}



\paragraph{Software}

Code is in Python and uses
PySINDy~\cite{deSilva2020,Kaptanoglu2022}, 
PySR~\cite{Cranmer23}, and 
ODEFormer~\cite{ascoli2024odeformer}.

\paragraph{Benchmarks}

DYSTS~\cite{gilpin2021chaos} and ODEBench~\cite{ascoli2024odeformer} with a total of 198 systems.

\paragraph{Foundation models}

We use \texttt{qwen2-72b-32k} for text tasks and \texttt{internvl2.5-78b} for vision-language tasks.
Temperature used for summarization is 1.0, otherwise 0.7.

\subsection{One-step of the framework}
\label{sec:expr:one-step}

Our initial experiment applies a single step of our framework without RAG. 
The prompt includes the main context and the summarized system description.
Ablations are defined as:
``None'' means no system description; 
``Text'' includes only $\mathcal{T}$; 
``Data'' includes only $\mathcal{D}$; and 
``Image'' includes only $\mathcal{I}$.
We generate 30 samples from the LLM, and select the one with the highest $R2$ score on test data.

\begin{figure}
    \centering
    \subfigure[Train]{
        \includegraphics[width=0.225\textwidth, trim={0.6cm 0.15cm 1cm 1.1cm}, clip]{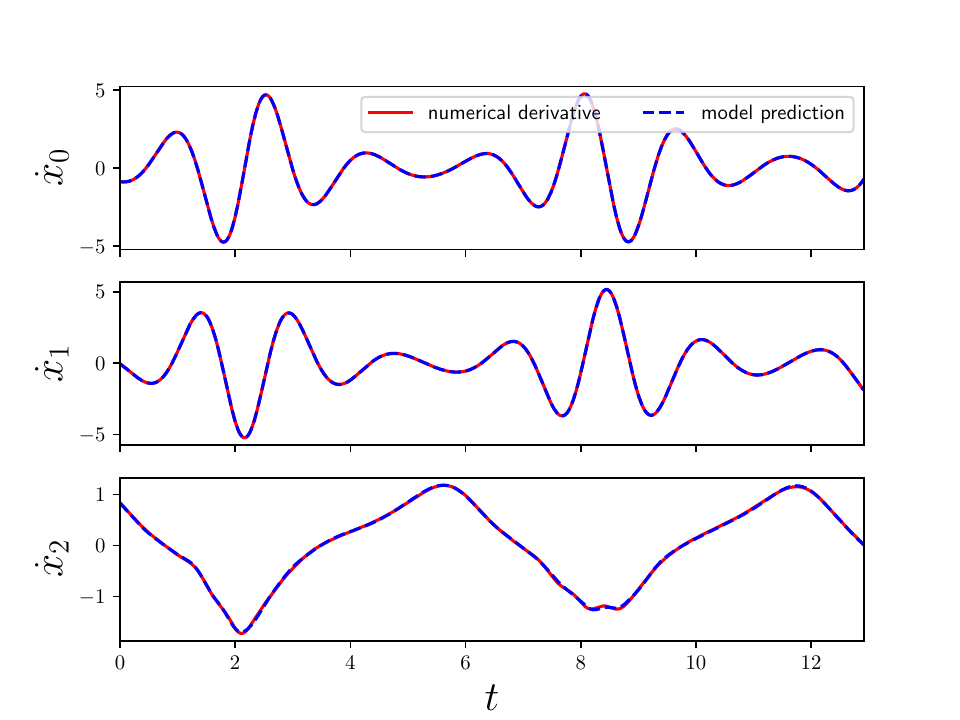}
        \label{fig:train-success-aizawa}
    }
    \hfill
    \subfigure[Test]{
        \includegraphics[width=0.225\textwidth, trim={0.6cm 0.15cm 1cm 1.1cm}, clip]{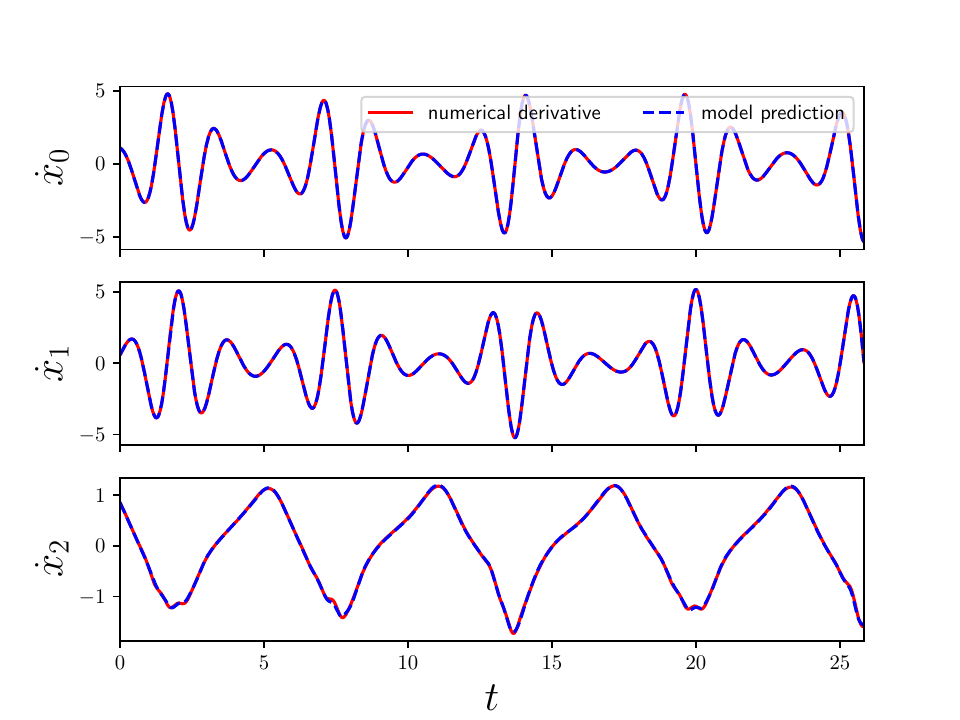}
        \label{fig:test-success-aizawa}
    }
    \caption{A successful trial where our framework produced a well-fitting model that generalized ($R2>0.99$ on training/test data), showing the numerical derivative~$\dot{x}$ with the SINDy model.}
    \label{fig:success-aizawa}
\end{figure}

An example of a well-fitting model on both the training and test data is shown in Figure~\ref{fig:success-aizawa}. 
The results for this step are reported in Table~\ref{tab:benchmark}, specifically in the rows labeled ``Ours'' with the respective ablation (``N'', ``T'', ``D'', ``I''). 
We observe, even a single step  of our method without RAG examples, is capable of identifying well-fitting models that generalize effectively to the test data.

\subsection{Examining Retrieval-Augmented Generation}

We look at the impact of including RAG examples into the prompt. 
Using results from Section~\ref{sec:expr:one-step}, we extracted the systems that failed to achieve $R2 > 0.99$ on the test data in each of the ablation.

We assume $\mathcal{T}$ is given, in order to use in the RAG retrieval (see~Section~\ref{sec:rag}).
We found  Nomic Embed~\cite{nussbaum2024nomicembedtrainingreproducible} effective as a text embedding model; 
CLIP~\cite{radford2021learning} was also tested but had insufficient context length.
Thus, we use the following ablations: ``Text'', ``Data'', and ``Image''.


 

\begin{table*}[]
\centering\tiny
\caption{Percentage of models with positive $PI$ ($\%, PI > 0$) across ablations ($N$ models per ablation) for two benchmarks. Additionally, the percentage of models exceeding thresholds (0.9 and 0.99) on training and test datasets is reported for $R=1, 5, 10$.}\label{tab:rag}
\begin{tabular}{|ll|c|ccc|cccccc|cccccc|}
\hline
                                                &   &     & \multicolumn{3}{c|}{\multirow{2}{*}{$\%,PI>0$}} & \multicolumn{6}{c|}{$R2 > 0.9$}                                     & \multicolumn{6}{c|}{$R2 > 0.99$}                                    \\ \cline{7-18} 
                                                &   &     & \multicolumn{3}{c|}{}                                & \multicolumn{3}{c|}{Train}              & \multicolumn{3}{c|}{Test} & \multicolumn{3}{c|}{Train}              & \multicolumn{3}{c|}{Test} \\
                                                &   & $N$ & 1                & 5               & 10              & 1    & 5    & \multicolumn{1}{c|}{10}   & 1       & 5      & 10     & 1    & 5    & \multicolumn{1}{c|}{10}   & 1      & 5       & 10     \\ \hline
\multicolumn{1}{|l|}{\multirow{3}{*}{\rotatebox{90}{DYSTS}}}    & T & 34  & 38.2             & 38.2            & 50.0            & 52.9 & 52.9 & \multicolumn{1}{c|}{50.0} & 35.3    & 35.3   & 35.3   & 23.5 & 23.5 & \multicolumn{1}{c|}{20.6} & 2.9    & 2.9     & 8.8    \\
\multicolumn{1}{|l|}{}                          & D & 37  & 59.5             & 59.5            & 62.2            & 51.4 & 54.1 & \multicolumn{1}{c|}{54.1} & 35.1    & 40.5   & 35.1   & 29.7 & 35.1 & \multicolumn{1}{c|}{32.4} & 8.1    & 13.5    & 10.8   \\
\multicolumn{1}{|l|}{}                          & I & 35  & 48.6             & 60.0            & 60.0            & 51.4 & 60.0 & \multicolumn{1}{c|}{54.3} & 22.9    & 28.6   & 31.4   & 31.4 & 31.4 & \multicolumn{1}{c|}{31.4} & 5.7    & 2.9     & 5.7    \\ \hline
\multicolumn{1}{|l|}{\multirow{3}{*}{\rotatebox{90}{ODEB.}}} & T & 24  & 41.7             & 41.7            & 37.5            & 62.5 & 66.7 & \multicolumn{1}{c|}{58.3} & 20.8    & 20.8   & 20.8   & 45.8 & 50.0 & \multicolumn{1}{c|}{50.}  & 8.3    & 4.2     & 4.2    \\
\multicolumn{1}{|l|}{}                          & D & 26  & 30.8             & 30.8            & 34.6            & 61.5 & 69.2 & \multicolumn{1}{c|}{80.8} & 19.2    & 23.1   & 26.9   & 42.3 & 42.3 & \multicolumn{1}{c|}{42.3} & 3.8    & 3.8     & 0.0    \\
\multicolumn{1}{|l|}{}                          & I & 27  & 29.6             & 48.1            & 33.3            & 70.4 & 74.1 & \multicolumn{1}{c|}{77.8} & 18.5    & 29.6   & 29.6   & 48.1 & 51.9 & \multicolumn{1}{c|}{55.6} & 3.7    & 11.1    & 3.7    \\ \hline
\end{tabular}
\end{table*}

The results of this experiment are shown in Table~\ref{tab:rag}. 
Overall, our findings suggest that RAG can yield positive outcomes. 
In most cases, performance correlates positively with the number of RAG examples used. 
However, $R2$ scores on the test data are generally weak, indicating that while RAG helps identify reasonable function approximators for the training data, these approximations do not necessarily generalize well.

\subsection{Improving the model with reflection}

Here, we examine the benefits of \textit{reflection} (i.e, utilizing previous attempts). 
Using results from Section~\ref{sec:expr:one-step}, we identified systems that failed to achieve $R2>0.99$ on the test data. 
For each system, we applied reflection, treating the one-step results as a first iteration. 
The ablations used (``None'', ``Text'', ``Data'', ``Image'') are as defined in Section~\ref{sec:expr:one-step}.
Each iteration, generates 30 samples and selects the one with the highest $R2$ score on the test data. 
This is compared with the previous best, and the highest $R2$ score is used as the candidate for that iteration.
Termination occurs when $R2>0.99$ on the test data. 
Again, the LLM is tasked only with selecting the feature library.

\begin{figure*}
    \includegraphics[width=\linewidth]{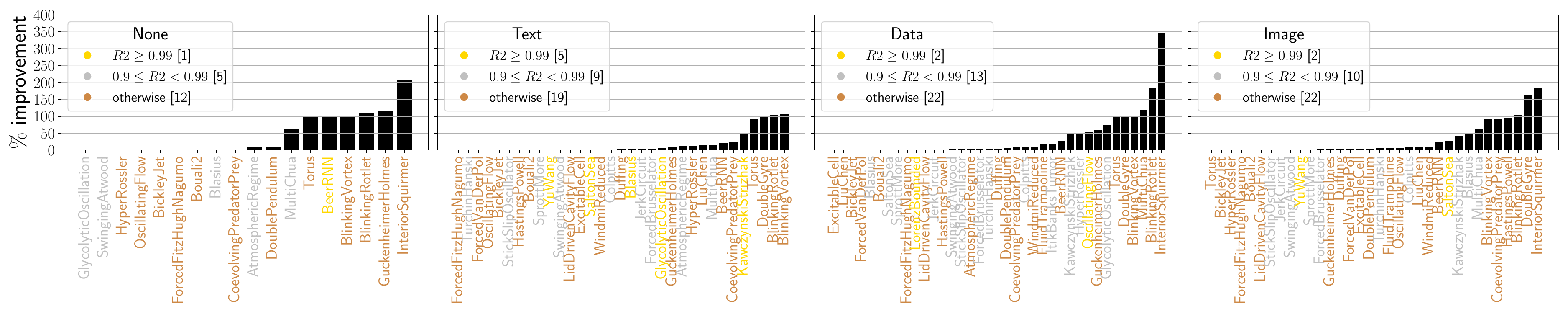}\\
    \includegraphics[width=\linewidth]{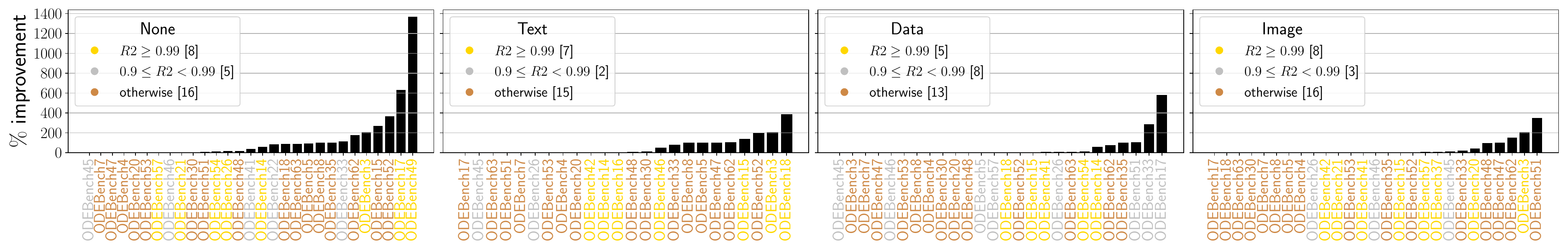}
    \caption{Percentage improvement in the $R2$ score after 10 iterations of reflection on the test dataset for the DYSTS (top row) and ODEBench (bottom row) benchmarks. Colors indicate whether the final $R2$ score is above a threshold. The numbers in $[\cdot]$ are the number of systems in this category.}
    \label{fig:perc-improve-reflection}
\end{figure*}

Results are summarized in Figure~\ref{fig:perc-improve-reflection}, that shows the $\%$ improvement after 10 iterations on the test dataset. 
Colors indicate whether a model achieved a final $R2$ score above a threshold: gold for $R2 \geq 0.99$, silver for $0.9 \leq R2 < 0.99$, and bronze otherwise. 
Additional plots are provided in the Appendix.

We see that reflection leads to significant improvements, up to approximately $350\%$ on the DYSTS benchmark and nearly $1400\%$ on ODEBench. 
Interestingly, even without a system description, high performance improvements are still achieved, highlighting the LLM's reasoning capabilities in analyzing previous attempts and identifying improvements.

\subsection{Comparison against alternatives}

We compared our approach against several alternatives from the literature on both benchmarks mentioned in Section~\ref{sec:preliminaries}. 
(i) \textbf{PySINDy}~\cite{deSilva2020}: basic SINDy method with default polynomial feature library (degree 2) and an optimizer implementing sequentially thresholded least squares~\cite{Brunton2016}.  
(ii) \textbf{LLM4ED}~\cite{Mengge24}: symbolic regression using an LLM to generate expressions, optimize parameters, and refine selection via reflection.  
(iii) \textbf{PySR}~\cite{Cranmer23}: symbolic regression utilizing genetic programming.  
(iv) \textbf{ODEFormer}~\cite{ascoli2024odeformer}: pre-trained transformer-based that maps measurement data to candidate expressions.  
All methods were evaluated with default hyperparameters.
For ODEFormer, we tested with beam size 50 (their experiments showed this to lead to higher performance) and beam temperatures 0.1 and 0.5; we report results for beam temperature 0.5 since this lead to higher results.

As of submission of this article, the code for LLM4ED~\cite{Mengge24} had not been publicly released\footnote{\cite{Mengge24} provide the link: \href{https://github.com/menggedu/EDL}{https://github.com/menggedu/EDL}, which remains empty at the time of writing.}, preventing direct comparison on both benchmarks. 
LLM4ED reports results on a subset of ODEBench (16 models), while our experiments cover the full benchmark (63 models). 
To enable comparison, we extracted results for this subset and also report findings for the other alternative methods.

For each benchmark model, we applied our method and each alternative, training and evaluating them using the R2 score (see Section~\ref{sec:scoring}) on both training and test data. 
Table~\ref{tab:benchmark} reports the percentage of models exceeding two performance thresholds: 0.9 (indicating a reasonably good fit) and 0.99 (indicating a very close fit). 
High values on training data suggest effective function approximation, while high values on test data indicate better generalization. 
A high test percentage may also suggest recovery of the original model, though this is not conclusive.

For our framework, we report both the overall results and three ablations. 
Each ablation corresponds to a scenario where only one step of the framework is applied, omitting reflection. 
In these ablation cases, no RAG examples were provided in the prompt, and the system description included either nothing (N), text (T), data summarization (D), or image summarization (I). 
Additionally, for the ablations, the optimizer was not selected—only the feature library was determined by the LLM. 
The overall results include all experimental conditions, incorporating RAG examples, all three system description components, reflection, human feedback, and optimizer selection.
Additionally, on particularly hard systems, we ran reflection with all three components incorporated into the system description; this helped to improve the overall performance.

\begin{table}[t]
\centering\tiny
\caption{Percentage of models with $R2$ score greater than a threshold (0.9 and 0.99) for the training and test data.}
\label{tab:benchmark}
\def\colwidth{0.26cm}  
\setlength{\tabcolsep}{3.75pt} 
\begin{tabular}{|>{\raggedright\arraybackslash}p{1cm}|*{4}{p{\colwidth}}|*{4}{p{\colwidth}}|*{4}{p{\colwidth}}|}
\hline
 & \multicolumn{4}{c|}{DYSTS} & \multicolumn{4}{c|}{ODEBench} & \multicolumn{4}{c|}{ODEBench (subset)}\\
\cline{2-13} 
& \multicolumn{2}{c|}{$>0.9$} & \multicolumn{2}{c|}{$>0.99$} & \multicolumn{2}{c|}{$>0.9$} & \multicolumn{2}{c|}{$>0.99$} & \multicolumn{2}{c|}{$>0.9$} & \multicolumn{2}{c|}{$>0.99$}  \\
Method & \rotatebox{90}{Train} & \multicolumn{1}{l|}{\rotatebox{90}{Test}} & \rotatebox{90}{Train} & \multicolumn{1}{l|}{\rotatebox{90}{Test}} & \rotatebox{90}{Train} & \multicolumn{1}{l|}{\rotatebox{90}{Test}} & \rotatebox{90}{Train} & \multicolumn{1}{l|}{\rotatebox{90}{Test}} & \rotatebox{90}{Train} & \multicolumn{1}{l|}{\rotatebox{90}{Test}} & \rotatebox{90}{Train} & \multicolumn{1}{l|}{\rotatebox{90}{Test}} \\
\cline{1-1} 
PySINDy      & 72.6 & \multicolumn{1}{l|}{64.4}  & 57.8              & \multicolumn{1}{l|}{54.1} & 66.7 & \multicolumn{1}{l|}{38.1} & 41.3 & 33.3 & 56.3 & \multicolumn{1}{l|}{31.3} & 31.3 & 31.3~~ \\
LLM4ED$^{*}$ & -    & \multicolumn{1}{l|}{-}     &         -         & \multicolumn{1}{l|}{-}    &    -  &    \multicolumn{1}{l|}{-}  &   -    & -  & \textbf{100} & \multicolumn{1}{l|}{\textbf{93.8}} & 93.8 & 75.0~~ \\
PySR         & 83.0 & \multicolumn{1}{l|}{73.3}  & 71.1              & \multicolumn{1}{l|}{64.4} & 93.7 & \multicolumn{1}{l|}{74.6} & 92.1 & 63.5 & \textbf{100} & \multicolumn{1}{l|}{81.3} & \textbf{100} & 68.8~~ \\
ODEFormer    & 0.7  & \multicolumn{1}{l|}{0.0}   & 0.0               & \multicolumn{1}{l|}{0.0} & 49.2 & \multicolumn{1}{l|}{34.9} & 33.3 & 25.4 & 62.5 & \multicolumn{1}{l|}{50.0} & 37.5 & 37.5 \\
Ours [N]     & 80.7 & \multicolumn{1}{l|}{72.59}  & 72.6              & \multicolumn{1}{l|}{65.2} & 82.5 & \multicolumn{1}{l|}{61.9} & 77.8 & 54.0 & 75.0 & \multicolumn{1}{l|}{43.8} & 62.5 & 37.5~~ \\
Ours [T]     & 85.2 & \multicolumn{1}{l|}{76.3}  & 75.6              & \multicolumn{1}{l|}{67.4} & 88.9 & \multicolumn{1}{l|}{69.8} & 77.8 & 61.9 & 81.3 & \multicolumn{1}{l|}{50.0} & 56.3 & 43.8~~ \\
Ours [D]     & 82.2 & \multicolumn{1}{l|}{71.1}  & 71.9              & \multicolumn{1}{l|}{64.4} & 88.9 & \multicolumn{1}{l|}{71.4} & 74.6 & 58.7 & 87.5 & \multicolumn{1}{l|}{68.8} & 74.6 & 58.7~~\\
Ours [I]     & 83.7 & \multicolumn{1}{l|}{72.6}  & 75.6              & \multicolumn{1}{l|}{67.4} & 88.9 & \multicolumn{1}{l|}{69.8} & 76.2 & 57.1 & 75.0 & \multicolumn{1}{l|}{56.3} & 75.0 & 43.8~~ \\
Ours [O]     & \textbf{94.8} & \multicolumn{1}{l|}{\textbf{84.4}} & \textbf{93.3} & \textbf{83.0} & \textbf{100} & \multicolumn{1}{l|}{\textbf{88.9}} & \textbf{98.4} & \textbf{85.7} & \textbf{100} & \multicolumn{1}{l|}{87.5} & \textbf{100} & \textbf{87.5}~~ \\
\hline
\end{tabular}
\begin{flushleft}
$^{*}$Results taken from~\cite{Mengge24}.
\end{flushleft}

\end{table}

Overall, our method outperforms alternatives in most cases, matches all instances where alternatives achieve 100\%, and falls slightly short in only one case. 
These results indicate that our approach is more effective at identifying function approximators that generalize well to unseen data.

\section{Discussion}

Overall, we have shown that our framework is highly capable at discovering physical laws from data; on two benchmarks, we have shown high performance.
In this section, we discuss several points of interest observed in the experiments. 

We conducted experiments incorporating RAG examples. 
While results show some improvement with RAG, and increasing the number of examples generally enhances performance, there are cases where omitting RAG entirely yields better outcomes than a single step of our framework with RAG. 
This may be due to prompt size increases, which have been observed to cause issues~\cite{fountas2024}. 
Additionally, the descriptions used in RAG used may not be ideal for finding examples that align in this context, suggesting that the query for generating the most similar examples might need refinement.

Human feedback has been explored in reflection-based approaches to enhance LLM decision-making, e.g.~\cite{ma2023eureka}. 
We conducted experiments to assess its impact on our framework. 
While human feedback provided some benefits, it generally did not yield significant improvements. 
Further details are available in the appendix.

In Table~\ref{tab:benchmark} in the subset of ODEBench experiments, the two examples where our method fails to achieve Test $R2 > 0.9$ are \texttt{ODEBench4} with $\dot{x} = \frac{1}{1 + e^{c_0 - x / c_1}}$ and \texttt{ODEBench7} with $\dot{x} = c_0x\log(c_1x)$. 
While our method successfully reproduces the training data, it struggles to generalize to test data for these cases. 
Upon inspecting these functions, the reason becomes evident: both are not expressed as a linear combination of primitive functions, thereby violating the core assumption of SINDy. 
Although our method finds a reasonable function approximator for the training data, this fundamental limitation of SINDy-based methods makes generalization difficult. 
Notably, symbolic regression can sometimes perform better, particularly in 1D systems. 
Interestingly, LLM4ED, which reports the discovered models in its paper, also encounters issues with these same two functions.

\section{Conclusion}

This paper presents a novel framework for discovering physical laws from data, leveraging large-scale pre-trained foundation models, as illustrated in Figure~\ref{fig:method-overview}. 
Our approach integrates multiple input modalities (text, raw data, and images) to enable model discovery. 
Building on the SINDy method, our framework enhances the model's capability to select suitable feature libraries and optimizers. 
We provide extensive experimental results on two benchmarks containing 198 models. 
We will provide open-source access to our code, prompts, and datasets, upon publication.

\subsection{Limitations}

Methods like ours and others~\cite{Mengge24, ma2024dreureka} leverage reflection, often resulting in lengthy prompts. 
Initially, we aimed to include the full raw data and several plots. 
However, this quickly exceeded the model's context length limits. 
We utilized both an LLM and VLM to summarize the data and images. 
Managing long context lengths remains a recognized challenge~\cite{fountas2024}, limiting the effectiveness of reflection-based approaches.

\subsection{Future work}

The discovery of physical laws has applications in control theory~\cite{Brunton16b, sahoo2018learning, kaiser2018sparse} and reinforcement learning~\cite{arora2022modelbased, zolman2024sindyrlinterpretableefficientmodelbased}. 
We aim to extend our framework to control and build on our previous ideas~\cite{mower2024optimal} to synthesize optimal controllers from natural language.

Although our approach performs well with pre-trained foundation models, these models were not explicitly trained for our specific use cases. 
A promising direction for future work is to leverage our agent-based framework, define a reward function—such as a linear combination of $R2$ scores on training and test datasets—and fine-tune the LLM using reinforcement learning.




Adapting our framework for cases where noise is present is a natural extension for handling more realistic scenarios.
In this case, for example, how to compute function derivatives become important often requiring smoothing~\cite{Chartrand11}.
We plan to incorporate these additional choices into our framework in future releases of our code.


\section*{Impact Statement}

This paper presents work whose goal is to advance the fields of Machine Learning, Physical Sciences, and Engineering.
There are many potential societal consequences of our work, none of which we feel must be specifically highlighted here.

\bibliography{main}
\bibliographystyle{icml2021}

\clearpage              
\onecolumn              
\appendix               

\section{Choosing a foundation model}\label{sec:choose-model}

\begin{figure}[H]
    \centering
    \includegraphics[width=0.5\linewidth, trim={4cm 1cm 2cm 2cm}, clip]{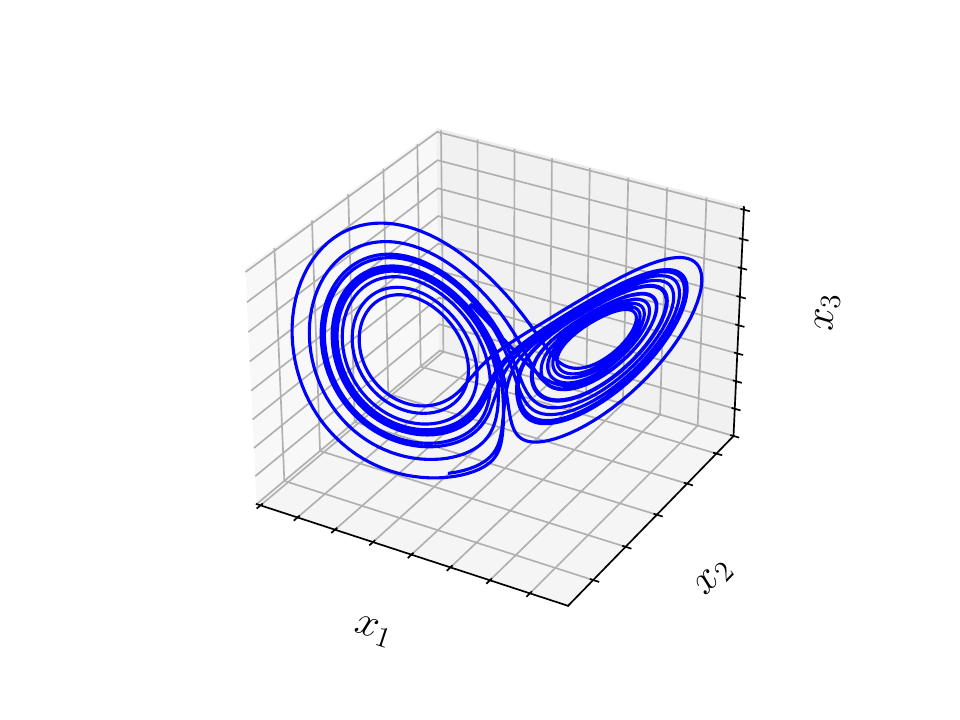}
    \caption{Example of the Lorenz system of equations.}
    \label{fig:lorenz}
\end{figure}

To promote reproducibility and transparency, we restrict ourselves to using only open-source models\footnote{Please note, the majority of this work was done prior to the release of DeepSeek-R1~\cite{deepseekai2025deepseekr1incentivizingreasoningcapability}.}.
Given the wide variety of pre-trained open-source models available, 
particularly from resources like HuggingFace, 
it can make it difficult to know which model is best to choose. 
To help guide this choice, we designed a brief experiment based on the Lorenz system of equations, shown in Figure~\ref{fig:lorenz}. 
Internally, Huawei hosts a collection of models hosted using vLLM~\cite{kwon2023efficient} available for continuous research use, from which we select models for this study.
This experiment also serves the secondary purpose of providing a controlled, simple test of our pipeline, allowing us to initially verify that our framework can reliably generate models, even in the case of a well-known system.

The Lorenz system consists of three nonlinear differential equations that model atmospheric convection and is famous for exhibiting chaotic behavior, where small changes in initial conditions can lead to drastically different outcomes.
These equations describe how variables related to temperature, fluid flow, and velocity evolve over time and are given by
\begin{equation}\label{eq:lorenz-system}
\begin{aligned}
\dot{x}_1 &= \sigma (x_2 - x_1) \\
\dot{x}_2 &= x_1 (\rho - x_3) - x_2 \\
\dot{x}_3 &= x_1x_2 - \beta x_3
\end{aligned}
\end{equation}
where $x = [x_1, x_2, x_3]^T\in\mathbb{R}^3$ represent the system's state variables (such as fluid velocity components), and $\sigma\in\mathbb{R}$, $\rho\in\mathbb{R}$, and $\beta\in\mathbb{R}$ are system parameters. 
In our experiments we use parameters $\sigma=10, \beta=2.66667, \rho=28$ which are common as default values in several libraries because they are associated with the original work by Edward Lorenz.
The Lorenz system is characterized by its sensitivity to initial conditions, which is a defining feature of chaotic behavior.

Given the model~\eqref{eq:lorenz-system} and an initial condition $x(0) = x_{init}$ and step size $\delta t$ (typically chosen to be small), we can use numerical integration to produce a trajectory $x(t)$ for $t\in[0, t_f]$.
We produce two trajectories, one that is used for training the model, and the second used for testing the trained model.
In our experiments we use $x_{init} = [-8, 8, 27]^T$ for training and $x_{init} = [8, 7, 15]^T$ for testing.
Both cases use $\delta t = 2\times 10^{-3}$.

\begin{figure}
    \centering
    \includegraphics[width=0.5\linewidth]{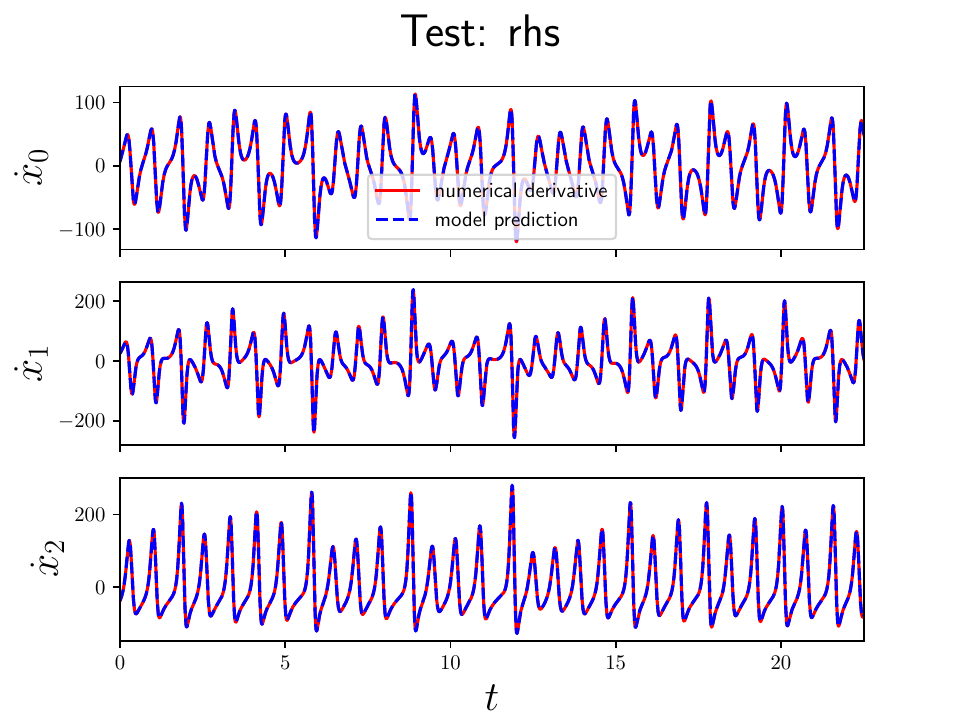}
    \caption{A successful fit for the Lorenz system of equations.}
    \label{fig:lorenz-fit}
\end{figure}

In this experiment we test both language and vision-language models and filter several viable models that we choose to use in the following experiments.
We tested 28 open-source models and consider a success when $R2\geq 0.99$.
The models that succeeded are shown in Table \ref{tab:best-foundation-models-lorenz} and an example of a successful fit is shown in Figure~\ref{fig:lorenz-fit}.

\begin{table}[H]
    \centering
    \scriptsize  
    \caption{Highest score for the Lorenz system for 23 models.}\label{tab:best-foundation-models-lorenz}
    \begin{tabular}{|c|c|c|c|}
    \hline
    Foundation Model & $R2$ & Foundation Model & $R2$\\
    \hline
    \texttt{qwen2.5-7b-instruct} & 0.9999999834634324 & \texttt{qwq-32b-preview-awq} & 0.9999999849116827 \\
    \texttt{qwen2.5-72b-32k} & 0.9999999849116827 & \texttt{qwen2.5-72b-instruct-lmdeploy} & 0.9999999849116827 \\
    \texttt{qwen2.5-32b-instruct-vllm} & 0.9999999849116827 & \texttt{internvl2.5-26b} & 0.9999999849116827 \\
    \texttt{qwen2.5-14b-instruct} & 0.9999999849116827 & \texttt{llama-3.1-nemotron-70b-instruct} & 0.9999999849116827 \\
    \texttt{codegeex4-all-9b} & 0.9999999849116827 & \texttt{qwen2.5-coder-32b-instruct} & 0.9999999849116827 \\
    \texttt{internvl2.5-38b} & 0.9999999849116827 & \texttt{qwen2.5-32b-instruct} & 0.9999999849116827 \\
    \texttt{qwen2.5-72b-instruct} & 0.9999999849116827 & \texttt{qwen2-72b-32k} & 0.9999999849116827 \\
    \texttt{functionary-small-v3.2} & 0.9999999849116827 & \texttt{llama-3.3-70b-instruct} & 0.9999999849116827 \\
    \texttt{qwen2-vl-72b-instruct} & 0.9999999849116827 & \texttt{qwq-32b-preview} & 0.9999999849116827 \\
    \texttt{llama-3.1-70b-instruct} & 0.9999999849116827 & \texttt{llama-3.1-8b-instruct} & 0.9999999849116827 \\
    \texttt{llama3.1-70b} & 0.9999999849116827 & \texttt{internvl2.5-78b} & 0.9999999849119878 \\
    \texttt{qwen2.5-coder-7b-instruct} & 0.9999999865971606 & & \\
    \hline
    \end{tabular}
\end{table}

Several models did not perform well and were not included in the potential models we use in later experiments.
The following notes are reported that provide a briefing on what the issue seemed to be. 
Please note, that these models may indeed see high performance in other tasks, and here we only report what we observed for this experiment.

\begin{enumerate}
    \item \texttt{qwen2.5-1.5b-instruct}: generated responses but nothing usable (e.g. undefined variables, un-imported classes).
    \item \texttt{minicpm-v-2.6}: generated responses but nothing usable (e.g frequent syntax errors).\
    \item \texttt{internvl2-40b}: responses received but very low quality (e.g. sometimes ``\texttt{``` python}'' (i.e. with an additional space) given instead of ``\texttt{```python}'', and other errors).
    \item \texttt{qwen2.5-3b-instruct}: generated responses but nothing usable (e.g. undefined variables, un-imported classes).
    \item \texttt{internvl2-26b}: generated responses but nothing usable (e.g. undefined variables, un-imported classes).
\end{enumerate}

Several successful models were filtered and reported in Table~\ref{tab:best-foundation-models-lorenz}.
Generally, increasing the number of model parameters during pre-training of large language models typically leads to better performance~\cite{Lan2020Albert}.
Thus, we chose to use the \texttt{qwen2-72b-32k} since it has been reported to have high performance on both coding and mathematics benchmarks~\cite{bai2023} and performed well in the authors own subjective experience.

\newpage

\section{Prompts}

In this section, we provide the prompts used in our experiments.
Note that some sections of the prompts have been removed for brevity.
You can find full unedited prompts used for our experiments in the supplementary material.

\subsection{Example main prompt}

\begin{prompt}[H]
\begin{tcolorbox}[promptbox, width=\textwidth]
\small
You are an expert at inferring dynamical systems using data-driven methods and we will use a method called Sparse Identification of Nonlinear Dynamics (SINDy).\\
*Requirements:*\\
* If given, analyze the system observation and identify relevant primitive functions.\\
* Effectively utilize available PySINDy feature libraries.\\
* Create a comprehensive feature library containing the primitive functions.\\
* Provide the library in a PySINDy-compatible format (no usage examples).\\
* Also, if given, analyze the system observation and identify a relevant optimizer for for the SINDy optimization process.\\
* Effectively utilize available optimizers, if one does not work well, try something else.\\
* If using NumPy, include the import statement.\\
* Import all feature libraries with from pysindy.feature\_library import * to avoid import errors.\\
*Chain of Thought:*\\
1. Understand the System: Review observations to grasp key components and behaviors.\\
2. Select Features: Choose feature libraries with appropriate arguments and consider a custom library with appropriate additional primitive functions to model the dynamics.\\
3. Select Optimizer: Choose an optimizer with appropriate arguments.\\
3. Build the Library: Assemble selected functions into a PySINDy-compatible feature library.\\
4. Validate and Refine: Ensure the library balances simplicity and accuracy for optimal interpretability.\\
*Example:*\\
Define the feature\_library variable.\\
\texttt{```python}\\
\texttt{from pysindy.feature\_library import *}\\
\texttt{from pysindy.optimizers import ConstrainedSR3, FROLS, SR3, SSR, STLSQ, TrappingSR3}\\
\texttt{import numpy as np}\\
\texttt{... additional imports and setup ...}\\
\texttt{feature\_library = ...}\\
\texttt{optimizer = ... choose optimizer from above}\\
\texttt{```}\\
*Feature library:*\\
Here are the available feature libraries in PySINDy.\\
Please carefully read the documentation to understand how to use them to meet your requirements.\\
\{FEATURE\_LIBRARY\_DOC\}\\
*Optimizer:*\\
Here are the available optimizers in PySINDy.\\
Please carefully read the documentation to understand how to use them to meet your requirements.\\
\{OPTIMIZER\_DOC\}\\
The following are previous attempts, please analyze them carefully and identify what leads to high R2 scores, particularly on the test dataset.\\
\{PREVIOUS\_ATTEMPTS\}\\
*System observation:*\\
**Text description:**\\
\{TEXT\_DESCRIPTION\}\\
**Data description:**\\
\{DATA\_SUMMARIZATION\}\\
**Image description:**\\
\{IMAGE\_SUMMARIZATION\}
\end{tcolorbox}
\caption{An example of the main prompt used in our framework.}
\label{prompt:main}
\end{prompt}

\subsection{Data summarization}

\begin{prompt}[H]
\begin{tcolorbox}[promptbox, width=\textwidth]
\small
You will be shown time-series data with {n} dimension{s}.
Read over it carefully and provide a comprehensive description of the data.\\
Make sure to include in your detailed description:\\
* The shape and common features of the trajectory.\\
* Whether noise seems to be present, or if the curve is smooth.\\
* Does the data resemble any known dynamical systems.\\
* Does any dimension of the data repeat, i.e. whether it seems to have a certain frequency or period.\\
* Does it look like there are any relationships between each of the state dimensions?\\
* If it repeats, try and provide an estimate of its amplitude and period from the plot.\\
* Anything additional you observe about the data that you think is relevant to form a complete description.\\
The data is as follows:\\
t,{xdims}\\
{data}\\
\end{tcolorbox}
\caption{This prompt is used to summarize time series data.}
\label{prompt:datasum}
\end{prompt}

\subsection{Image summarization}

\begin{prompt}[H]
\begin{tcolorbox}[promptbox, width=\textwidth]
\small
You will be shown an image of a time-series plot of measured data with {n} dimension{s}.
Look at it carefully and provide a comprehensive description of what you see.\\
Make sure to include in your detailed description:\\
* The shape and common features of the trajectory.\\
* Whether noise seems to be present, or if the curve is smooth.\\
* Does the data resemble any known dynamical systems.\\
* Does any dimension of the data repeat, i.e. whether it seems to have a certain frequency or period.\\
* Does it look like there are any relationships between each of the state dimensions?\\
* If it repeats, try and provide an estimate of its amplitude and period from the plot.\\
* Anything additional you observe about the data that you think is relevant to form a complete description.
\end{tcolorbox}
\caption{This prompt is used to summarize plots/images.}
\label{prompt:imagesum}
\end{prompt}

\newpage

\section{Output code example}

The following is an example output code from the LLM.
Minimal editing has been done to fit the code on the page.

{\small
\begin{verbatim}
from pysindy.feature_library import *
from pysindy.optimizers import ConstrainedSR3, FROLS, SR3, SSR, STLSQ, TrappingSR3
import numpy as np

# Polynomial terms to capture higher-order dynamics
poly_lib = PolynomialLibrary(degree=4, include_interaction=True, include_bias=True)

# Trigonometric terms to capture periodic oscillations
fourier_lib = FourierLibrary(n_frequencies=4)

# Custom library for exponential terms and logarithmic terms to fit potential 
# noise in the system
functions = [
    lambda x : np.exp(x), lambda x : np.exp(-x), lambda x : np.log(np.abs(x)), 
    lambda x : x**2, lambda x,y : x*y]
custom_exp_lib = CustomLibrary(library_functions=functions)

# Given that x0, x1, and x2 have different dynamics, we can consider custom libraries 
# for each
# For x0, which exhibits a trend, we may use polynomial terms and perhaps logarithmic 
# terms
# For x1 and x2, which show oscillatory behavior and potential noise, we can include 
# trigonometric functions and exponential terms
# Combine libraries
feature_library = GeneralizedLibrary([poly_lib, custom_exp_lib, fourier_lib])

# Choose SR3 optimizer with a suitable threshold
# The threshold can be tuned based on validation set performance
optimizer = SR3(threshold=0.03)
\end{verbatim}
}

\newpage

\section{One-step of the framework: additional results}\label{app:one-step-additional}

A single step of our framework failed to find models with $R2\geq 0.99$ and are summarized in Table \ref{tab:one-step-0-99-failures} below. 
These models are used in the later experiments, as we include additional ablations that examine if it is possible with other features to improve the performance.

\begin{table}[H]
    \centering
    \caption{Dynamical systems that failed to achieve $R2 \geq 0.99$. A total of 39 models are listed. A cross (\ding{55}) indicates that the model was not identified in a single step of our framework when the LLM only chose the feature library.}
    \label{tab:one-step-0-99-failures}
\begin{tabular}{|l|c|c|c|c|c|c|c|c|c|}
\hline
Dataset Name & None & Text & Data & Image & Dataset Name & None & Text & Data & Image \\
\hline
AtmosphericRegime & \ding{55} & \ding{55} & \ding{55} &   & HyperRossler & \ding{55} & \ding{55} & \ding{55} & \ding{55} \\ 
BeerRNN & \ding{55} & \ding{55} & \ding{55} & \ding{55} & InteriorSquirmer & \ding{55} &   & \ding{55} & \ding{55} \\ 
BickleyJet & \ding{55} & \ding{55} & \ding{55} & \ding{55} & ItikBanksTumor &   &   & \ding{55} &   \\ 
Blasius & \ding{55} & \ding{55} & \ding{55} & \ding{55} & JerkCircuit & \ding{55} & \ding{55} & \ding{55} & \ding{55} \\ 
BlinkingRotlet & \ding{55} & \ding{55} & \ding{55} & \ding{55} & KawczynskiStrizhak & \ding{55} & \ding{55} & \ding{55} & \ding{55} \\ 
BlinkingVortex & \ding{55} & \ding{55} & \ding{55} & \ding{55} & LidDrivenCavityFlow & \ding{55} & \ding{55} & \ding{55} & \ding{55} \\ 
Bouali2 & \ding{55} & \ding{55} & \ding{55} & \ding{55} & LiuChen & \ding{55} & \ding{55} & \ding{55} & \ding{55} \\ 
CoevolvingPredatorPrey & \ding{55} & \ding{55} & \ding{55} & \ding{55} & LorenzBounded &   &   & \ding{55} &   \\ 
Colpitts & \ding{55} & \ding{55} & \ding{55} & \ding{55} & MultiChua & \ding{55} & \ding{55} & \ding{55} & \ding{55} \\ 
DoubleGyre & \ding{55} & \ding{55} & \ding{55} & \ding{55} & OscillatingFlow & \ding{55} & \ding{55} & \ding{55} & \ding{55} \\ 
DoublePendulum & \ding{55} &   & \ding{55} & \ding{55} & SaltonSea & \ding{55} & \ding{55} & \ding{55} & \ding{55} \\ 
Duffing & \ding{55} & \ding{55} & \ding{55} & \ding{55} & SprottMore & \ding{55} & \ding{55} & \ding{55} & \ding{55} \\ 
ExcitableCell & \ding{55} & \ding{55} & \ding{55} & \ding{55} & StickSlipOscillator & \ding{55} & \ding{55} & \ding{55} & \ding{55} \\ 
FluidTrampoline & \ding{55} & \ding{55} & \ding{55} & \ding{55} & SwingingAtwood & \ding{55} & \ding{55} & \ding{55} & \ding{55} \\ 
ForcedBrusselator & \ding{55} & \ding{55} & \ding{55} & \ding{55} & Torus & \ding{55} & \ding{55} & \ding{55} & \ding{55} \\ 
ForcedFitzHughNagumo & \ding{55} & \ding{55} & \ding{55} & \ding{55} & TurchinHanski & \ding{55} & \ding{55} & \ding{55} & \ding{55} \\ 
ForcedVanDerPol & \ding{55} & \ding{55} & \ding{55} & \ding{55} & WindmiReduced & \ding{55} & \ding{55} & \ding{55} & \ding{55} \\ 
GlycolyticOscillation & \ding{55} & \ding{55} & \ding{55} & \ding{55} & YuWang & \ding{55} & \ding{55} &   & \ding{55} \\ 
GuckenheimerHolmes & \ding{55} & \ding{55} & \ding{55} & \ding{55} & SprottI & \ding{55} &   &   &   \\ 
HastingsPowell & \ding{55} & \ding{55} & \ding{55} & \ding{55} &  &  &  &  &  \\ 
\hline
\end{tabular}
\end{table}

\newpage

\section{Human feedback}\label{app:human-feedback}

This experiment examined the impact of human feedback on guiding the LLM to produce higher-performing models. 
At each iteration, participants reviewed the previously generated code and the trained model's performance on training and test data. 
They then provided input, which was incorporated into the next prompt alongside the prior code. 
To select models for each ablation, we identified failures from the initial experiment (Section~\ref{sec:expr:one-step}), filtering out models with $R2$ scores above 0.2 or below 0.99—ensuring a meaningful starting point. 
The remaining models were sorted in ascending order, and the five with the lowest test $R2$ scores were selected. 
For each of these models in each ablation, five iterations were conducted, with user feedback collected at each step.
Note, since there are different failure cases in the previous experiment, it is not necessarily the case that each of the five models in each ablation will be the same.

\begin{figure}[H]
    \includegraphics[width=\linewidth]{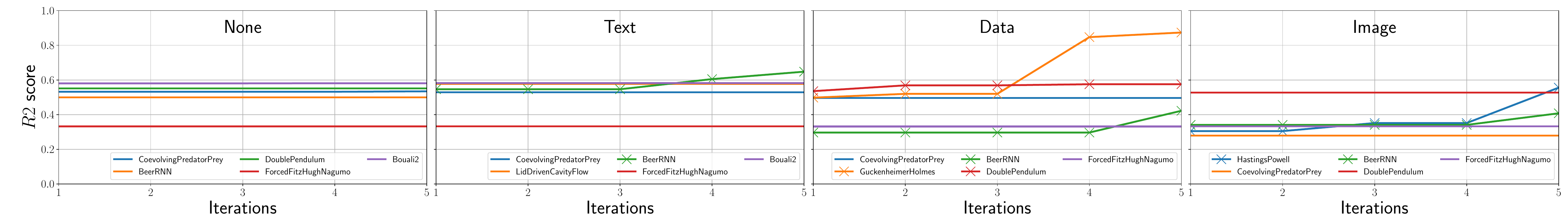}
    \caption{Human feedback can lead to improved $R2$ score.}
    \label{fig:human-feedback}
\end{figure}

The results of this experiment are presented in Figure~\ref{fig:human-feedback}; we only conducted this experiment using the DYSTS benchmark.
We observe that there are some cases where human feedback has helped improve the model fitting score, however, in many cases the performance was not improved.



\newpage

\section{Extended analysis: RAG}

The following images in this section show the percentage improvement for each system. 
The percentage in the title for each image indicates the percentage of the models that saw a positive improvement. 
A green bar indicates improvement, red indicates negative improvement, and black also indicates negative improvement but means the bar extends further than the limits for the graph. 
The percentage value rounded to one decimal place is also provided for each bar. 
Also, the model name is appended with a start $*$ when $R2 \geq 0.99$.

\subsection{DYSTS}

\begin{figure}[H]
    \centering
    \includegraphics[width=0.75\linewidth]{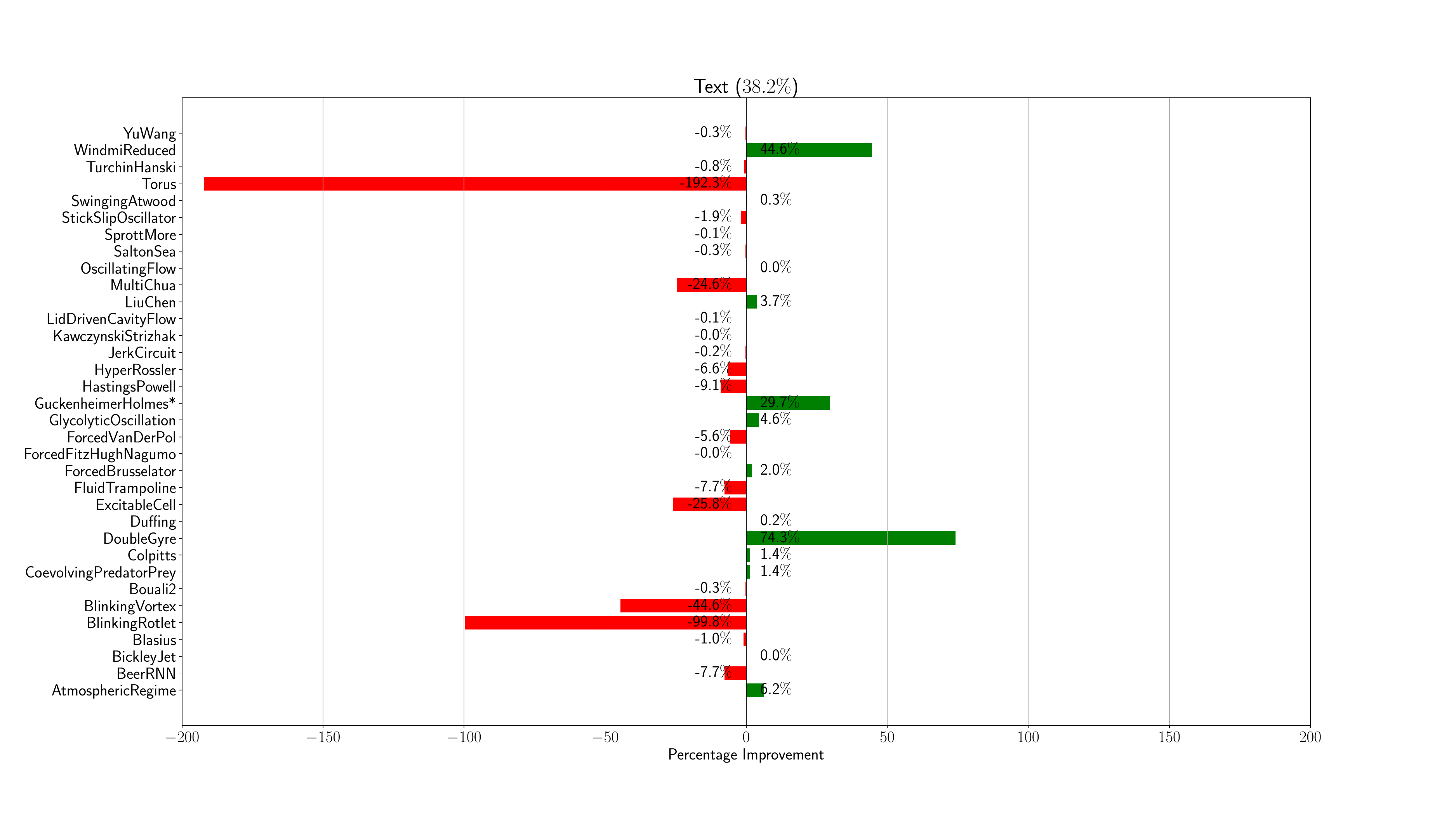}
    \caption{Percentage improvement for each system when using $R=1$ RAG examples with Text.}
    \label{fig:rag-t-1}
\end{figure}

\begin{figure}[H]
    \centering
    \includegraphics[width=0.75\linewidth]{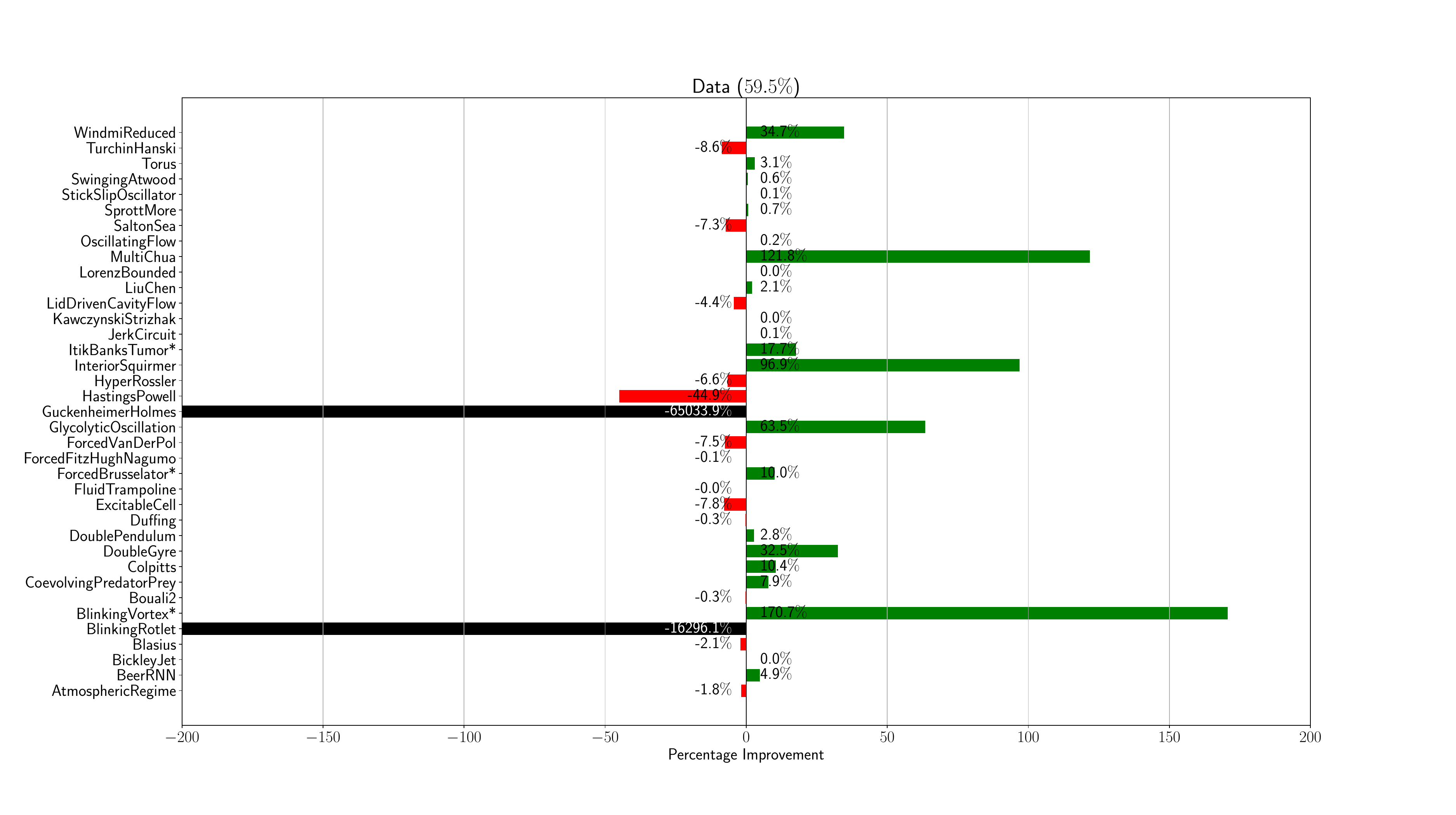}
    \caption{Percentage improvement for each system when using $R=1$ RAG examples with Data.}
    \label{fig:rag-d-1}
\end{figure}

\begin{figure}[H]
    \centering
    \includegraphics[width=0.75\linewidth]{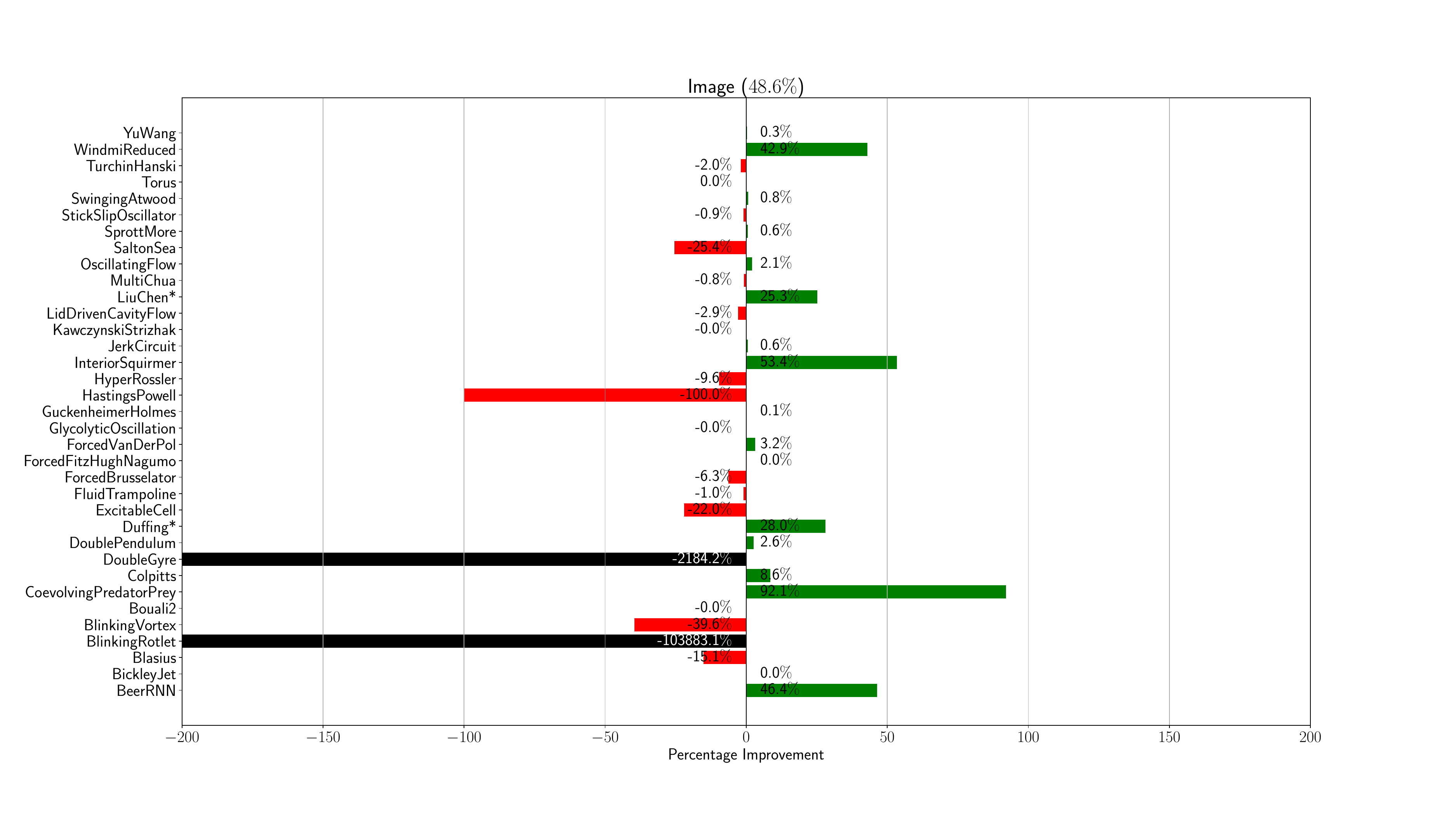}
    \caption{Percentage improvement for each system when using $R=1$ RAG examples with Image.}
    \label{fig:rag-i-1}
\end{figure}

\begin{figure}[H]
    \centering
    \includegraphics[width=0.75\linewidth]{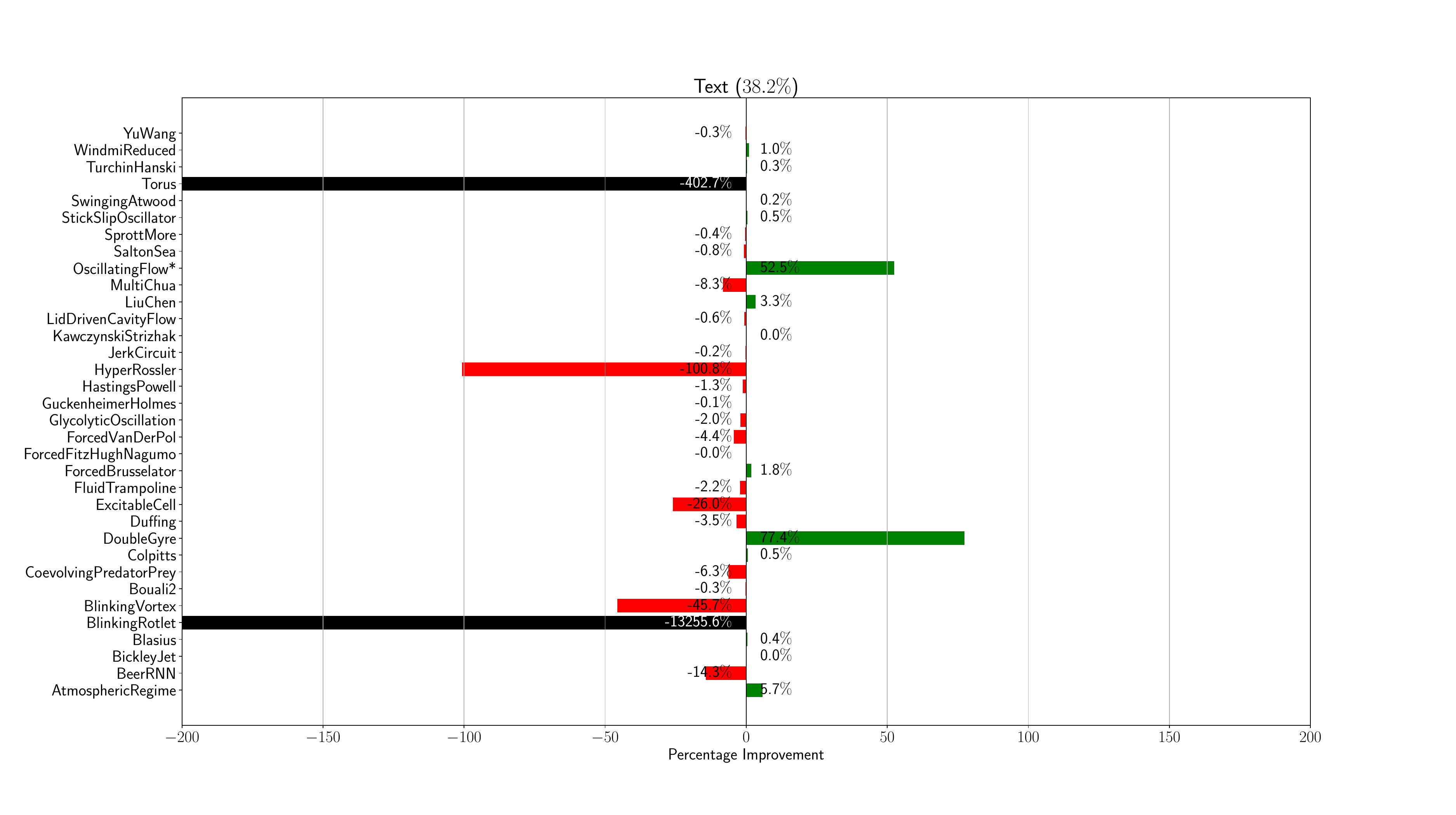}
    \caption{Percentage improvement for each system when using $R=5$ RAG examples with Text.}
    \label{fig:rag-t-5}
\end{figure}

\begin{figure}[H]
    \centering
    \includegraphics[width=0.75\linewidth]{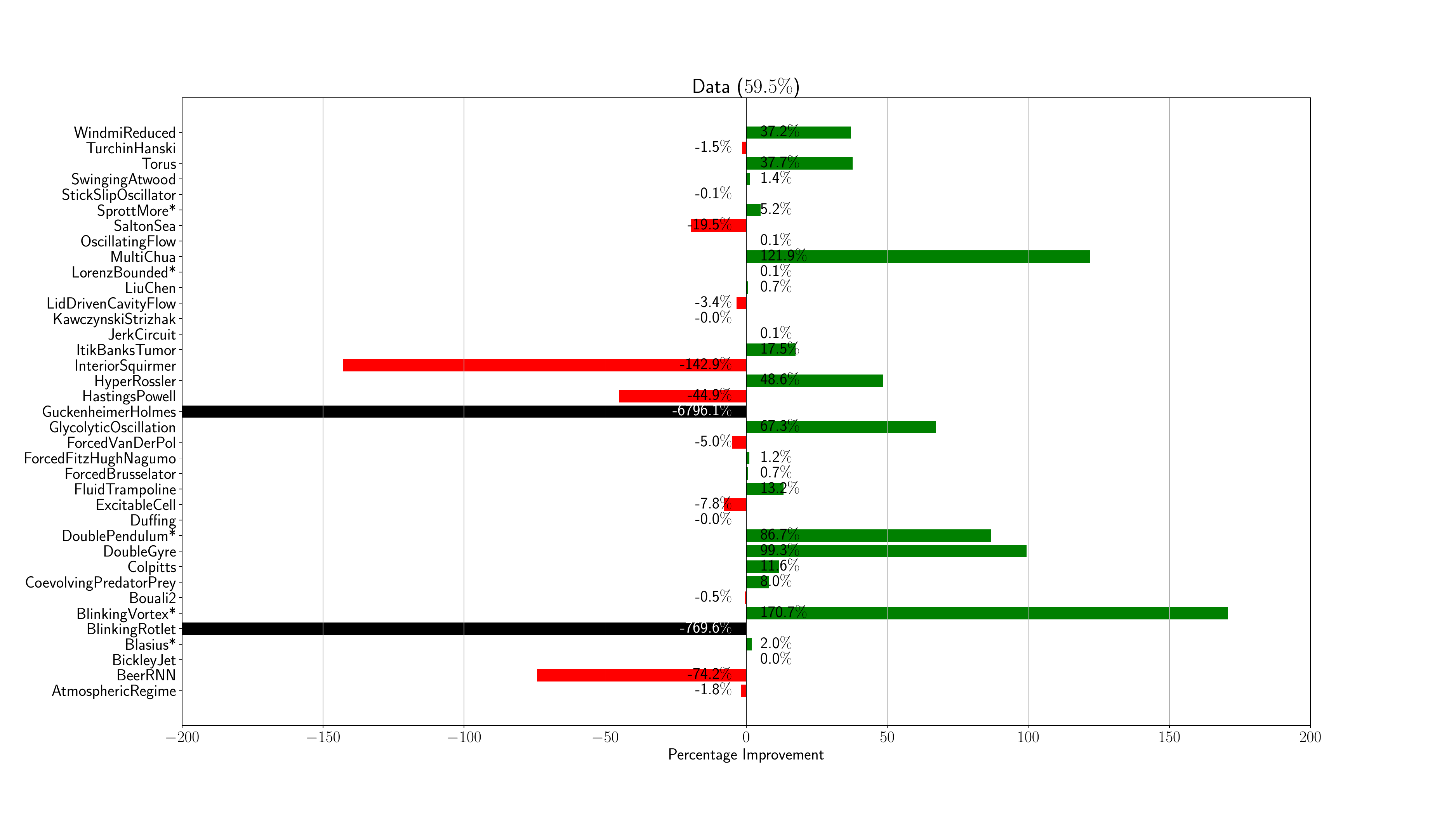}
    \caption{Percentage improvement for each system when using $R=5$ RAG examples with Data.}
    \label{fig:rag-d-5}
\end{figure}

\begin{figure}[H]
    \centering
    \includegraphics[width=0.75\linewidth]{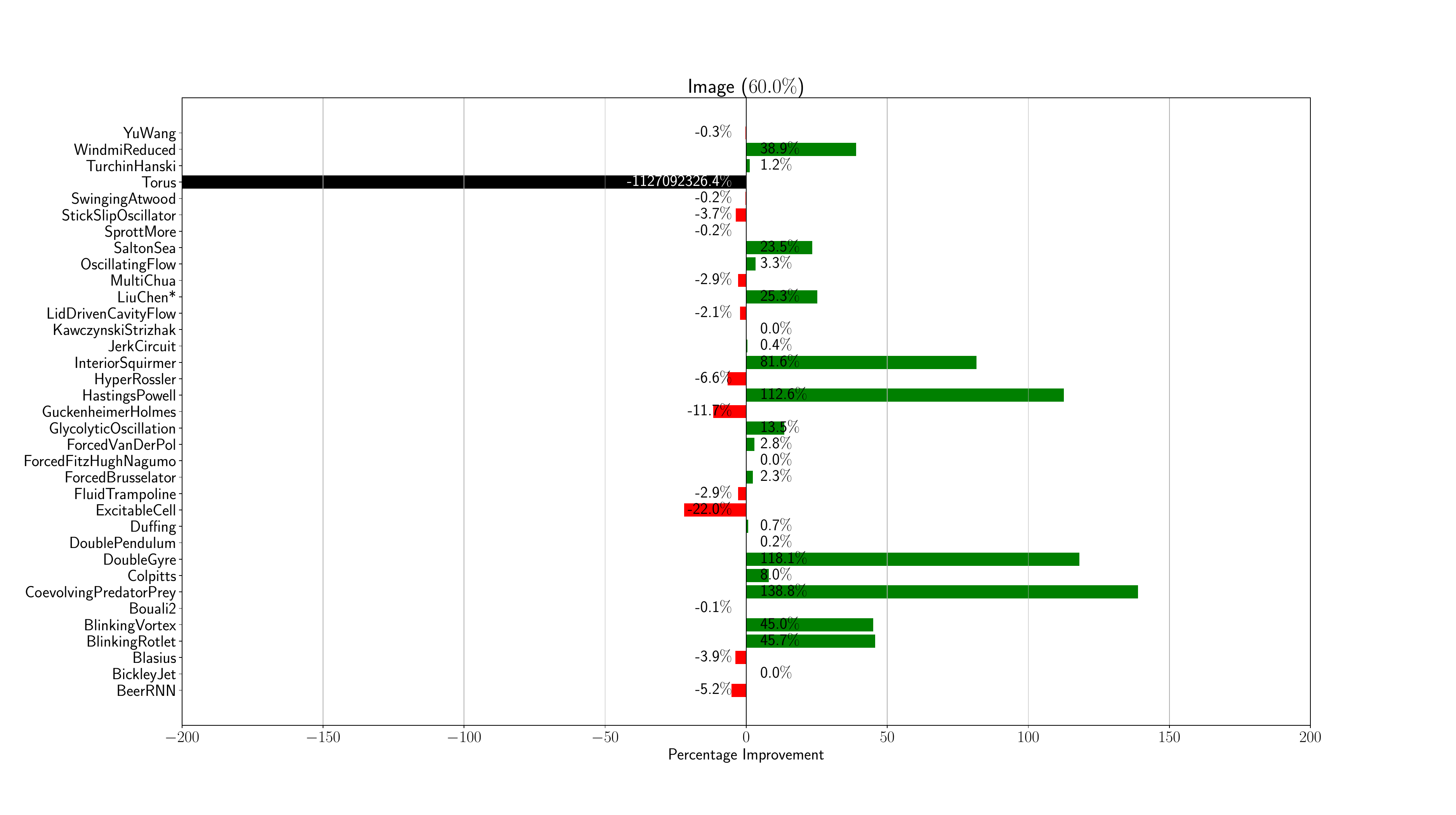}
    \caption{Percentage improvement for each system when using $R=5$ RAG examples with Image.}
    \label{fig:rag-i-5}
\end{figure}

\begin{figure}[H]
    \centering
    \includegraphics[width=0.75\linewidth]{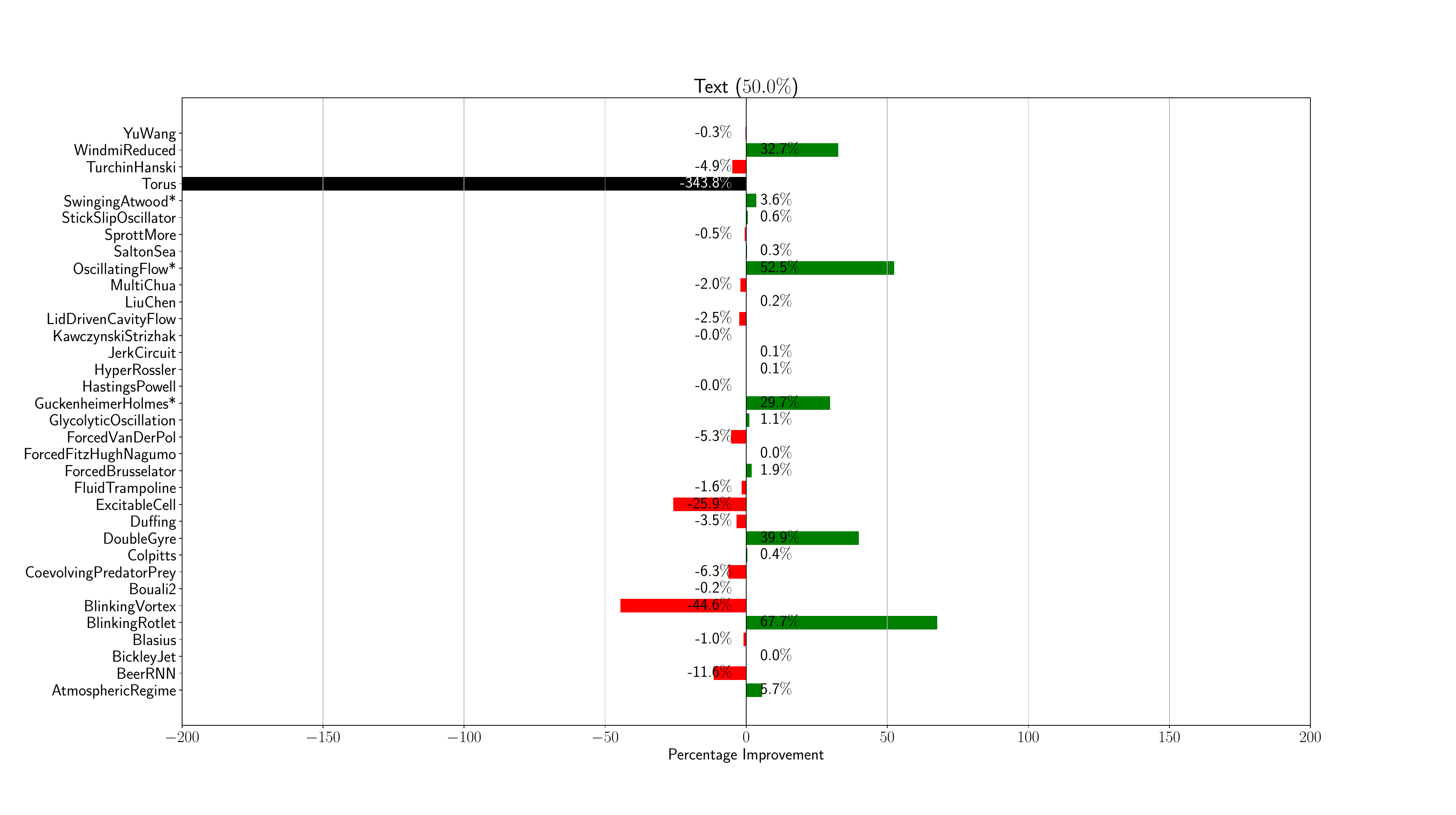}
    \caption{Percentage improvement for each system when using $R=10$ RAG examples with Text.}
    \label{fig:rag-t-10}
\end{figure}

\begin{figure}[H]
    \centering
    \includegraphics[width=0.75\linewidth]{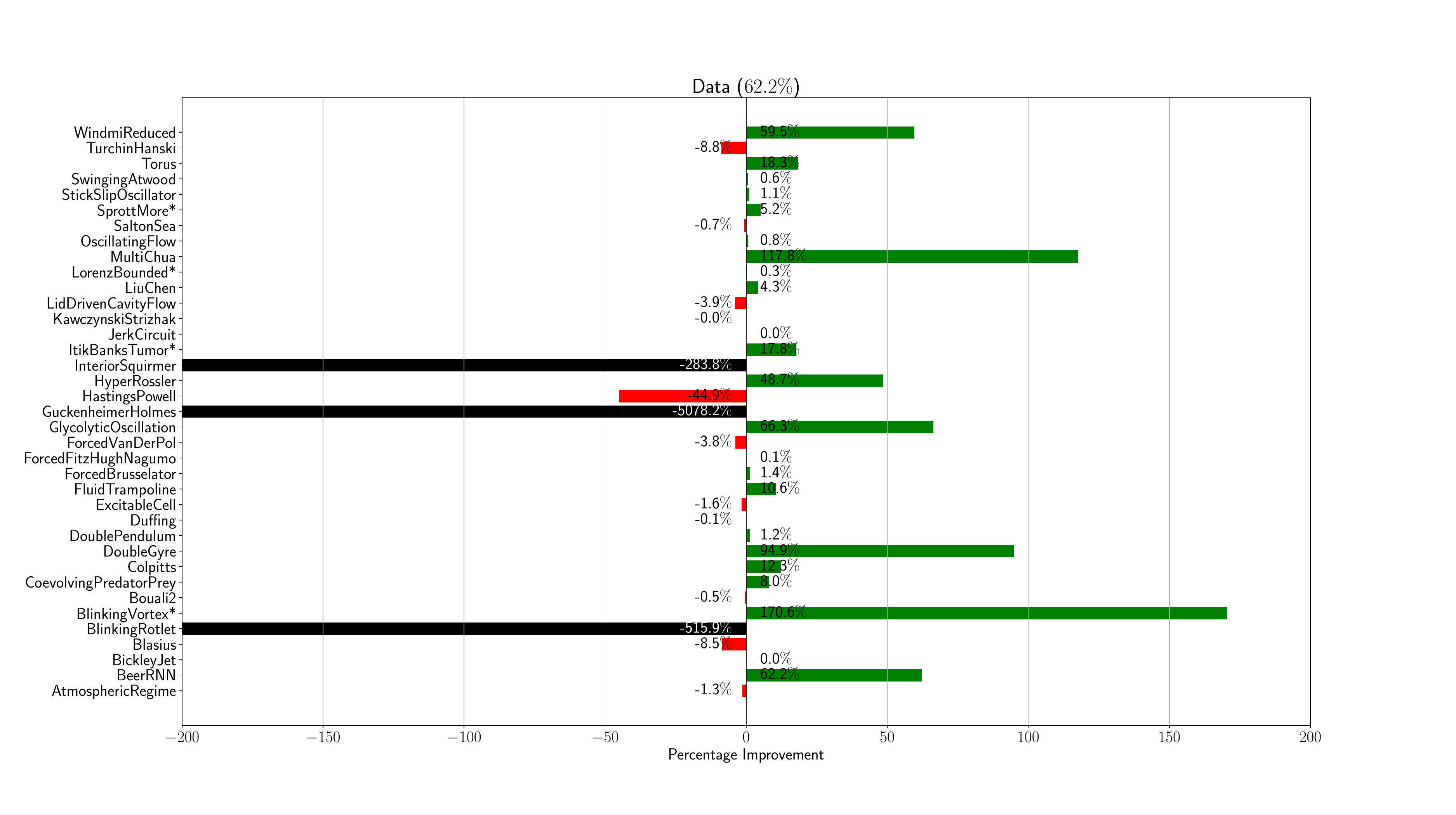}
    \caption{Percentage improvement for each system when using $R=10$ RAG examples with Data.}
    \label{fig:rag-d-10}
\end{figure}

\begin{figure}[H]
    \centering
    \includegraphics[width=0.75\linewidth]{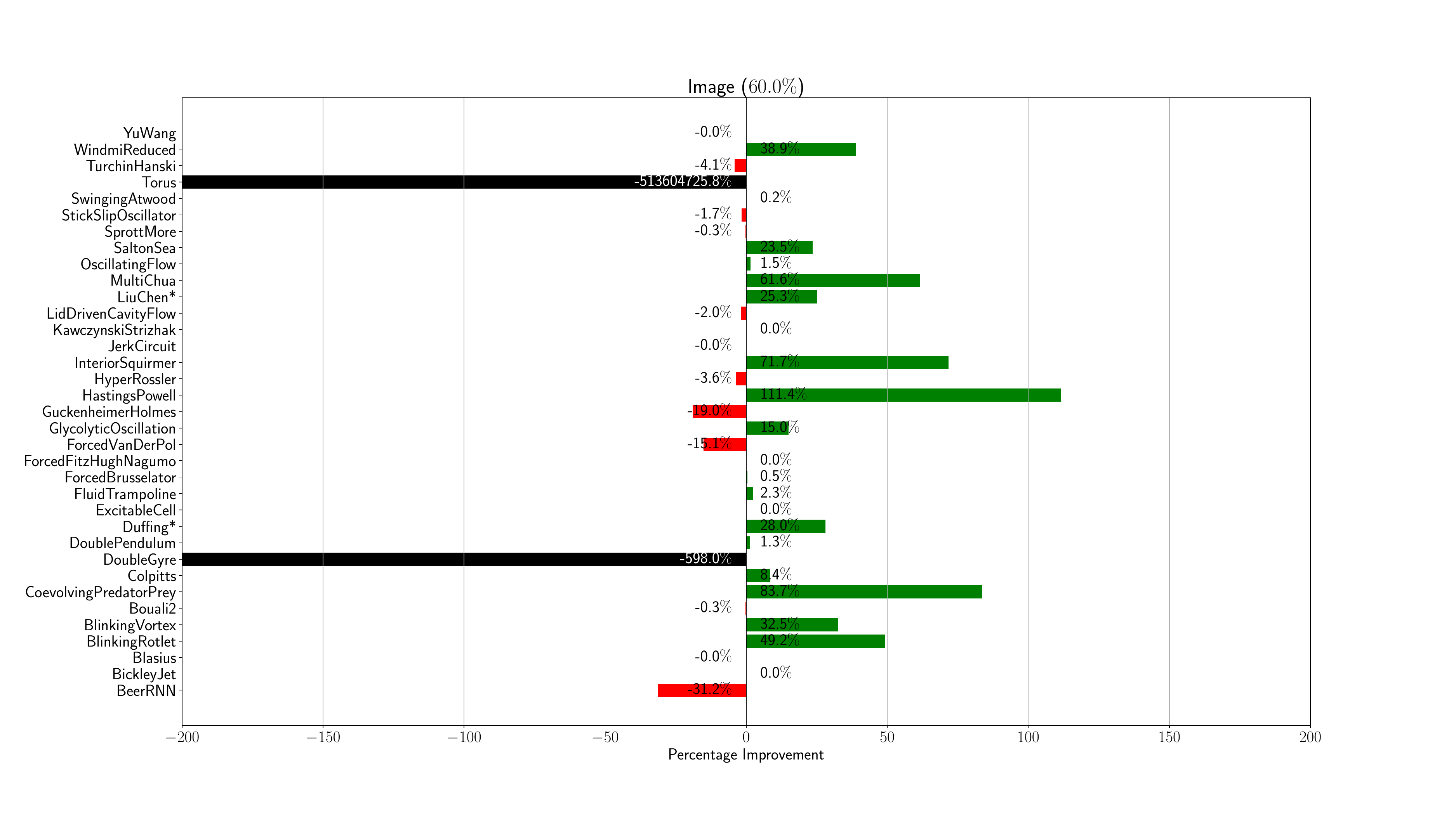}
    \caption{Percentage improvement for each system when using $R=10$ RAG examples with Image.}
    \label{fig:rag-i-10}
\end{figure}

\subsection{ODEBench}

\begin{figure}[H]
    \centering
    \includegraphics[width=0.75\linewidth]{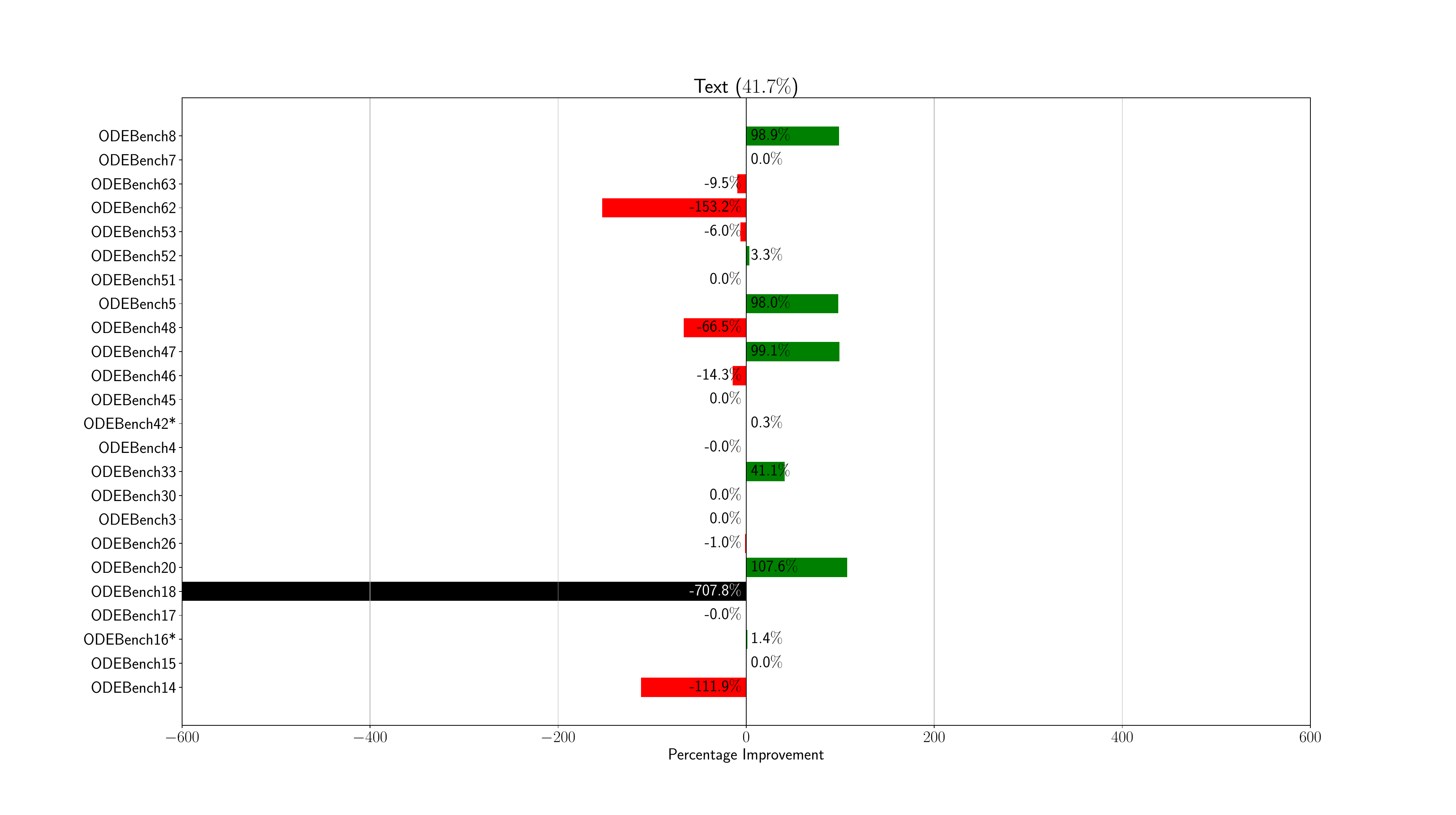}
    \caption{Percentage improvement for each system when using $R=1$ RAG examples with Text.}
    \label{fig:odebench_rag-t-1}
\end{figure}

\begin{figure}[H]
    \centering
    \includegraphics[width=0.75\linewidth]{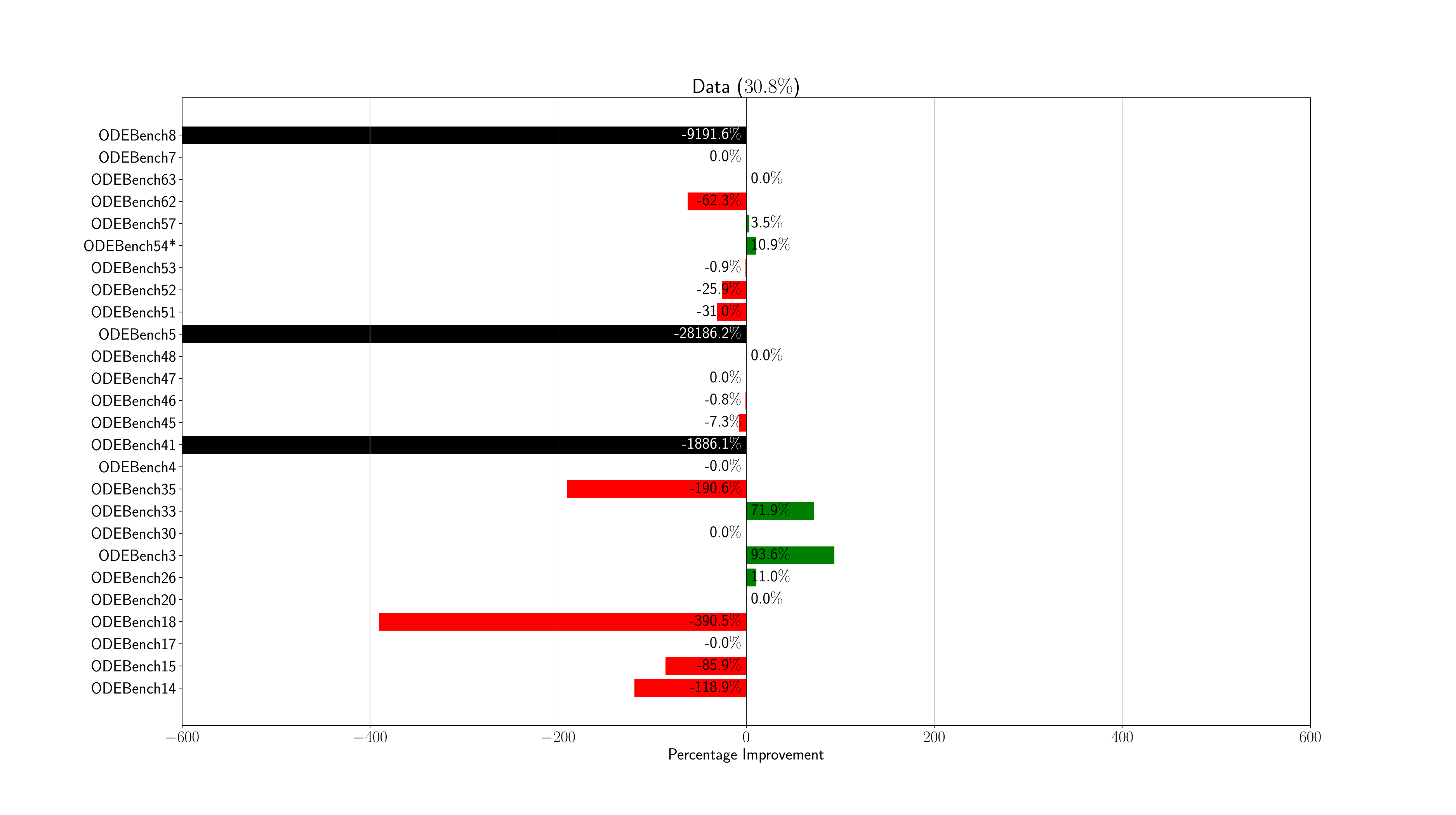}
    \caption{Percentage improvement for each system when using $R=1$ RAG examples with Data.}
    \label{fig:odebench_rag-d-1}
\end{figure}

\begin{figure}[H]
    \centering
    \includegraphics[width=0.75\linewidth]{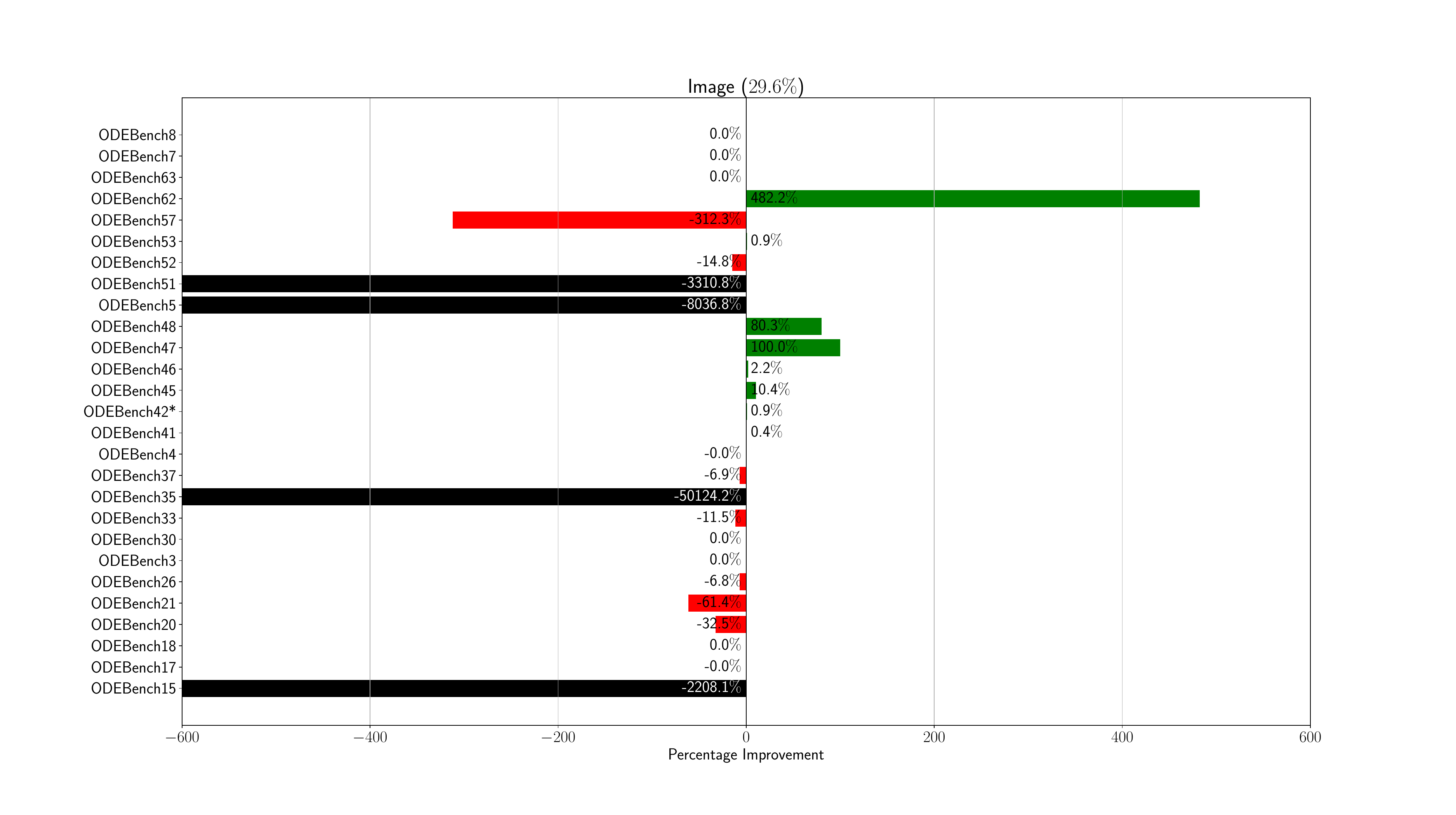}
    \caption{Percentage improvement for each system when using $R=1$ RAG examples with Image.}
    \label{fig:odebench_rag-i-1}
\end{figure}

\begin{figure}[H]
    \centering
    \includegraphics[width=0.75\linewidth]{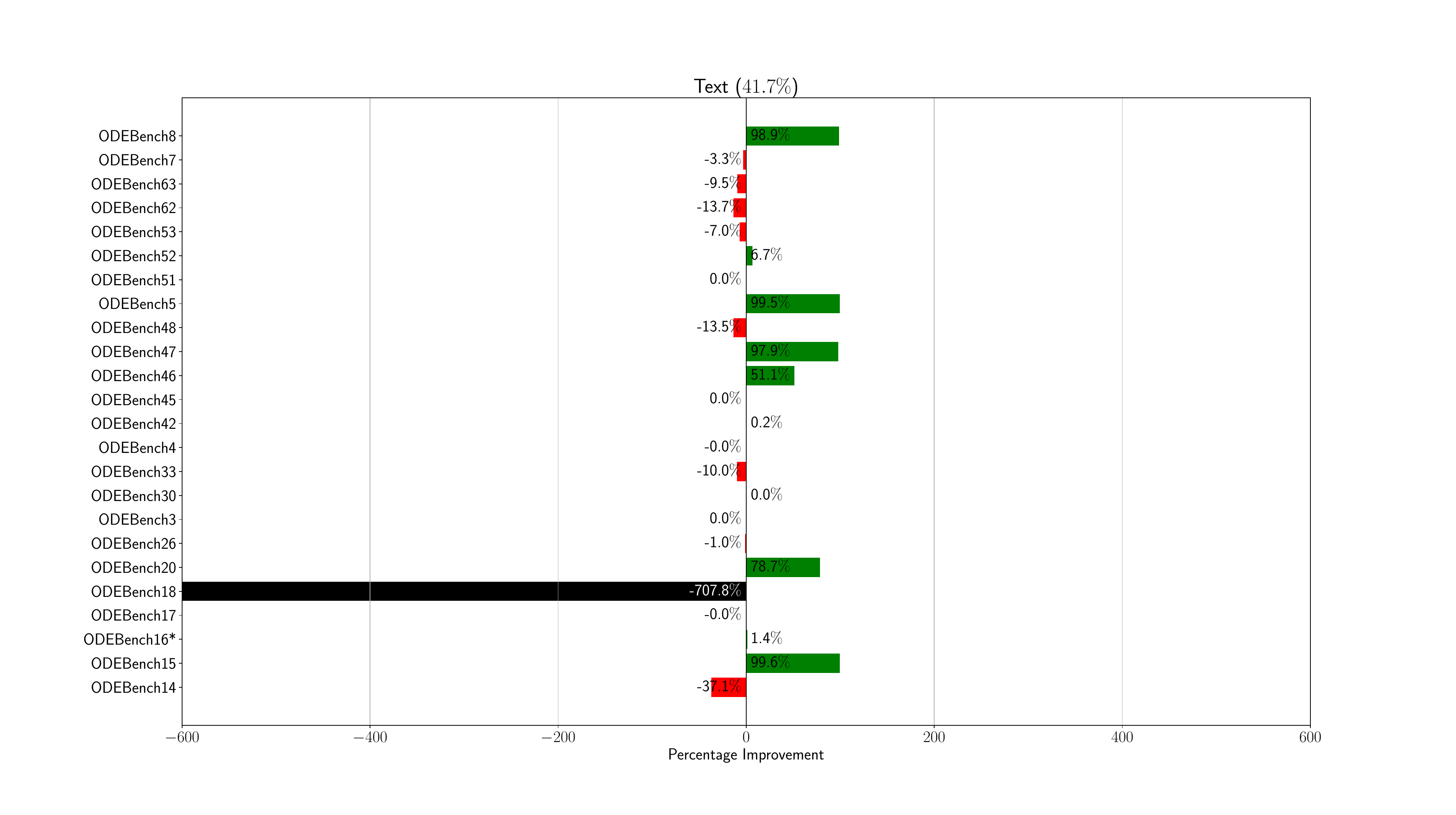}
    \caption{Percentage improvement for each system when using $R=5$ RAG examples with Text.}
    \label{fig:odebench_odebench_rag-t-5}
\end{figure}

\begin{figure}[H]
    \centering
    \includegraphics[width=0.75\linewidth]{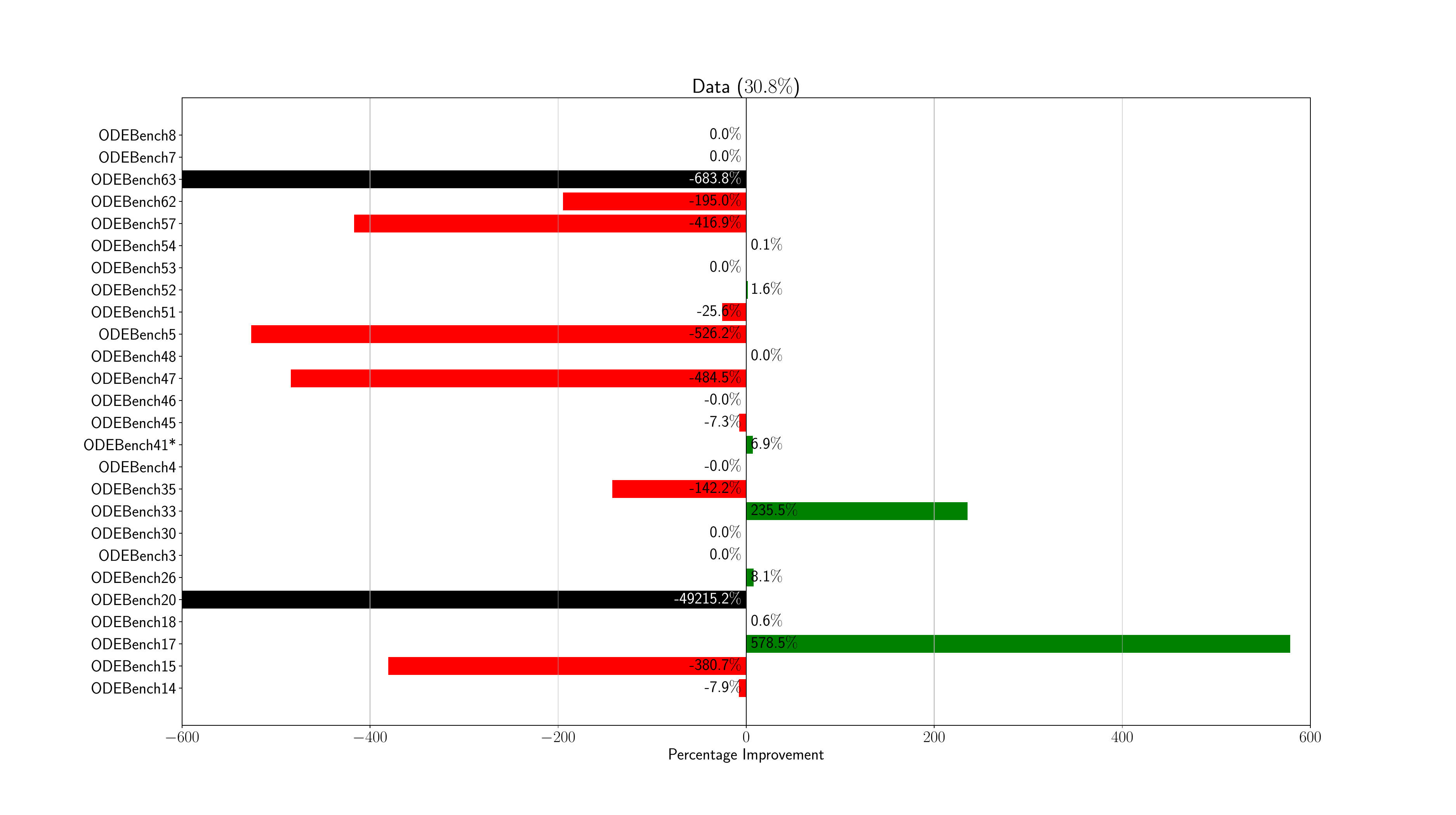}
    \caption{Percentage improvement for each system when using $R=5$ RAG examples with Data.}
    \label{fig:odebench_rag-d-5}
\end{figure}

\begin{figure}[H]
    \centering
    \includegraphics[width=0.75\linewidth]{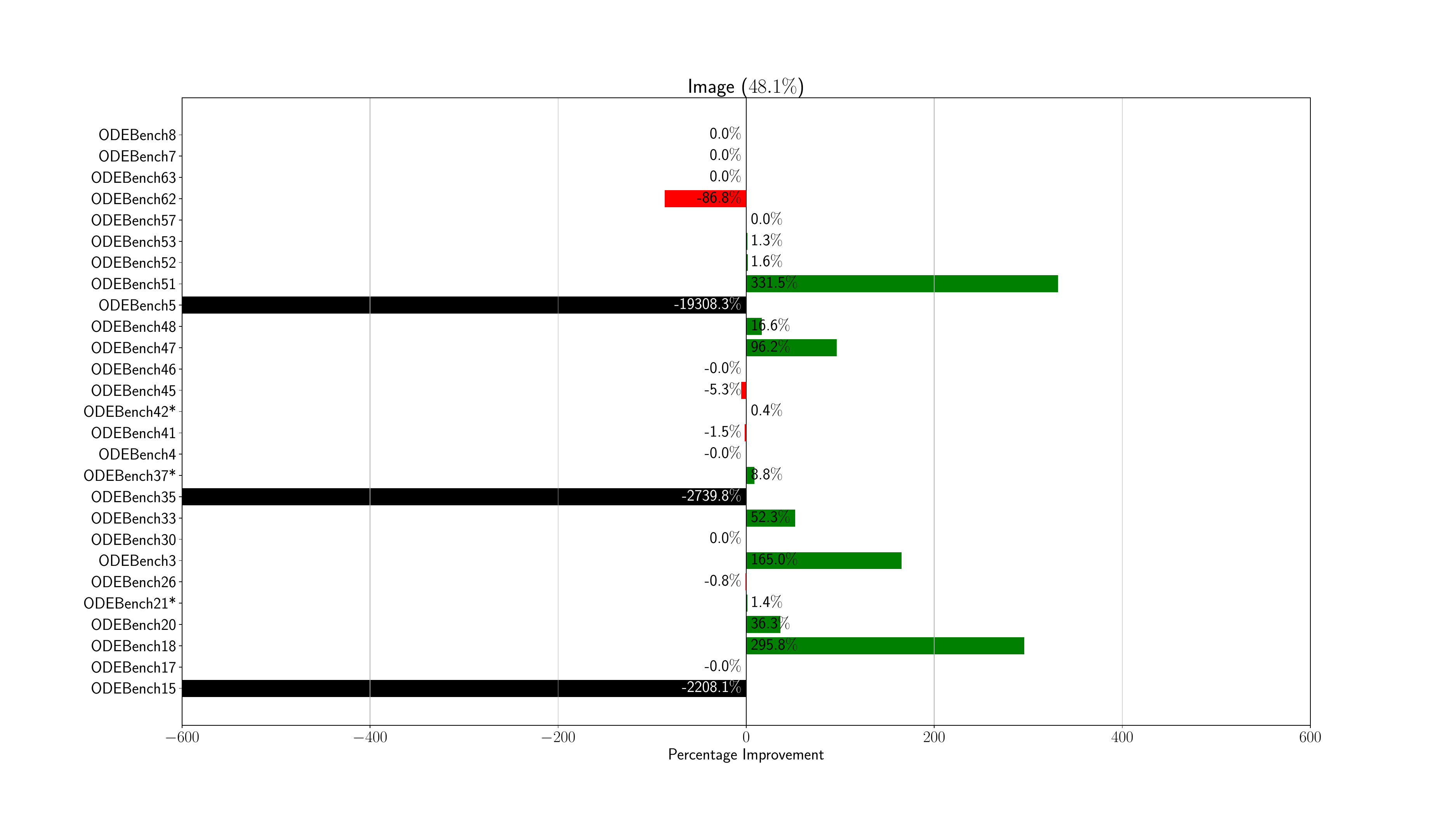}
    \caption{Percentage improvement for each system when using $R=5$ RAG examples with Image.}
    \label{fig:odebench_rag-i-5}
\end{figure}

\begin{figure}[H]
    \centering
    \includegraphics[width=0.75\linewidth]{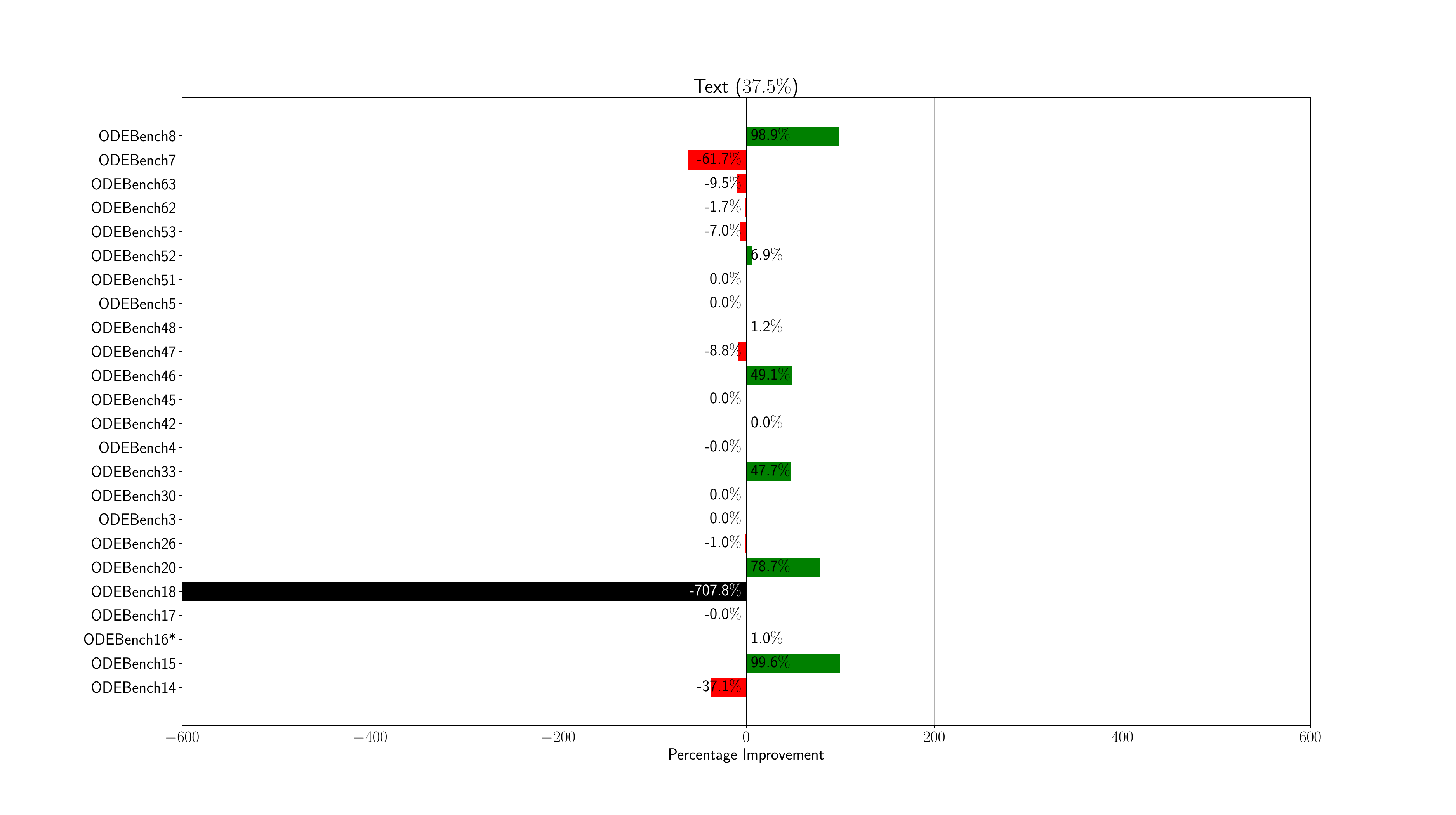}
    \caption{Percentage improvement for each system when using $R=10$ RAG examples with Text.}
    \label{fig:odebench_rag-t-10}
\end{figure}

\begin{figure}[H]
    \centering
    \includegraphics[width=0.75\linewidth]{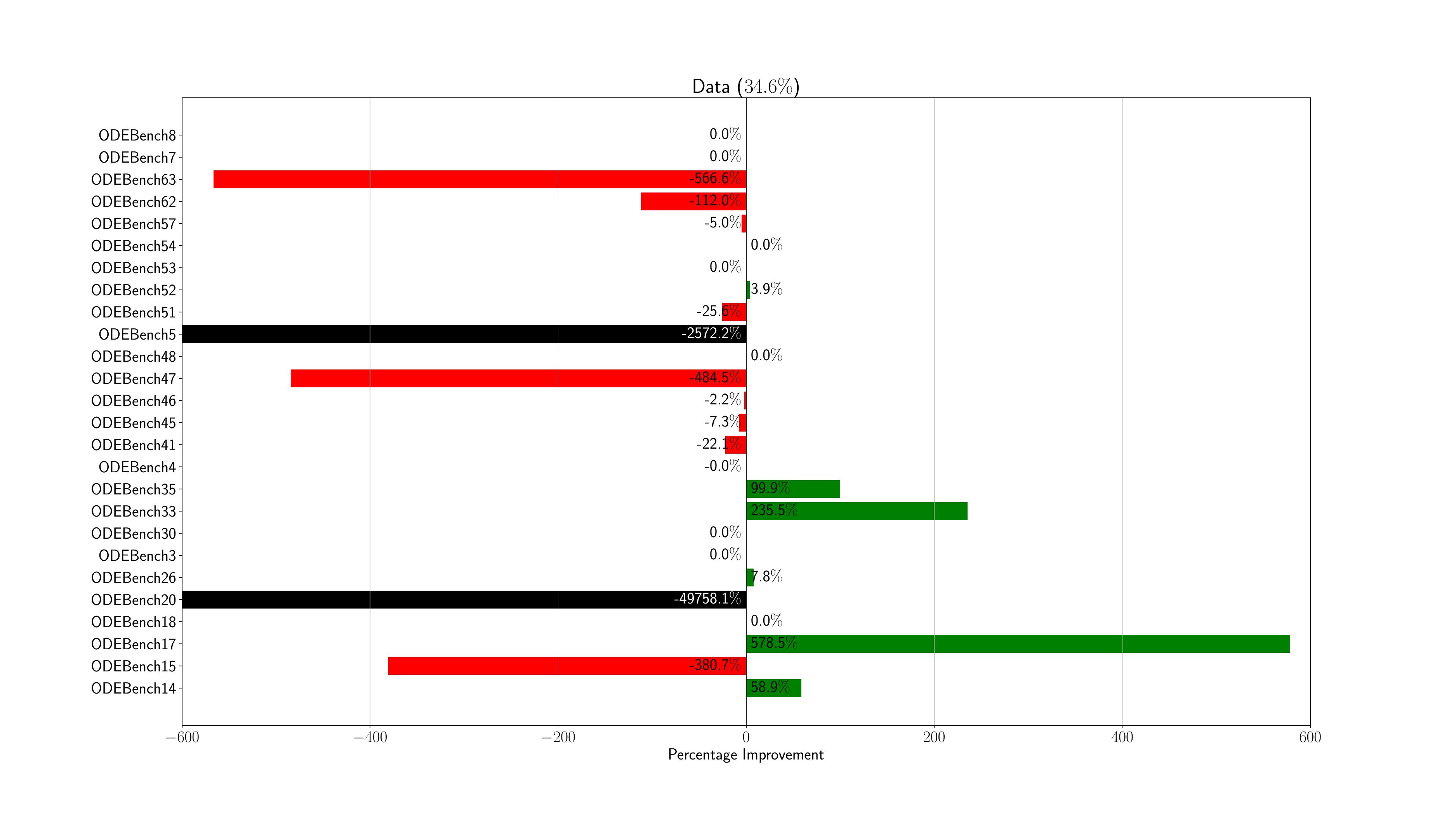}
    \caption{Percentage improvement for each system when using $R=10$ RAG examples with Data.}
    \label{fig:odebench_rag-d-10}
\end{figure}

\begin{figure}[H]
    \centering
    \includegraphics[width=0.75\linewidth]{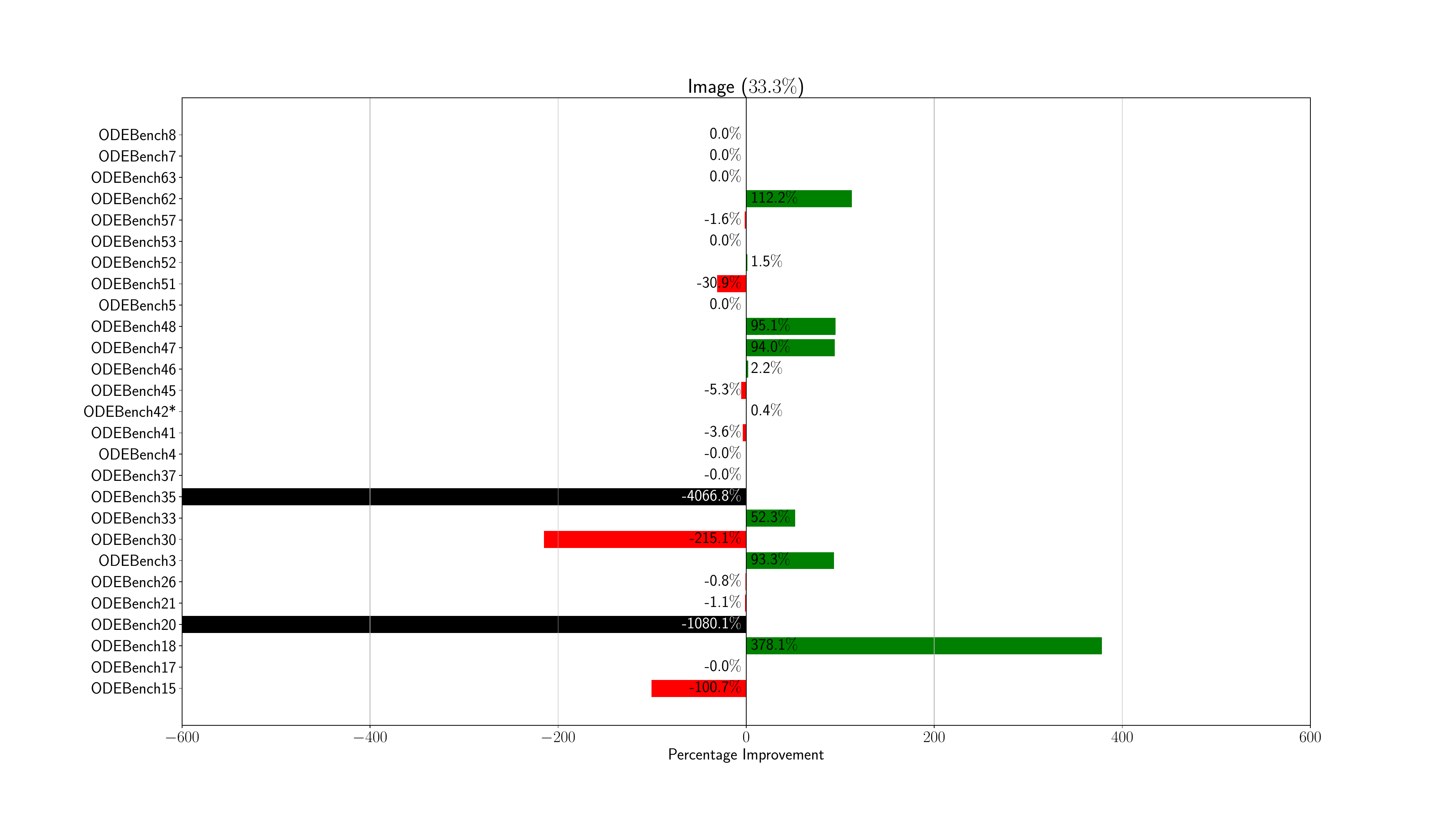}
    \caption{Percentage improvement for each system when using $R=10$ RAG examples with Image.}
    \label{fig:odebench_rag-i-10}
\end{figure}

\newpage

\section{Extended analysis: reflection}\label{app:reflection}

\begin{figure}[H]
    \centering
    \includegraphics[width=\linewidth]{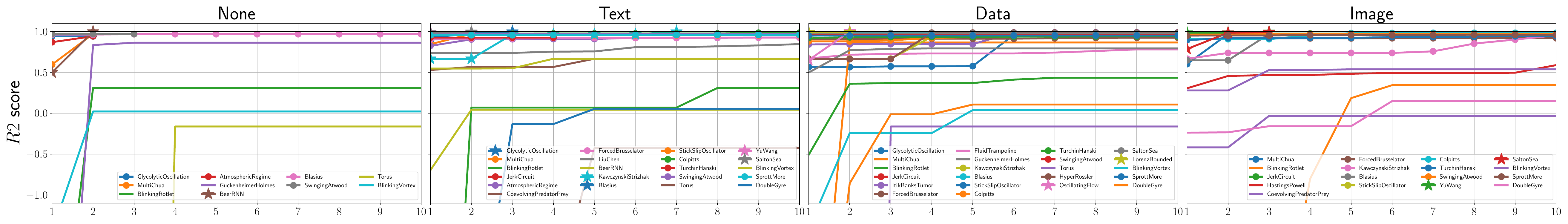}\\
    \includegraphics[width=\linewidth]{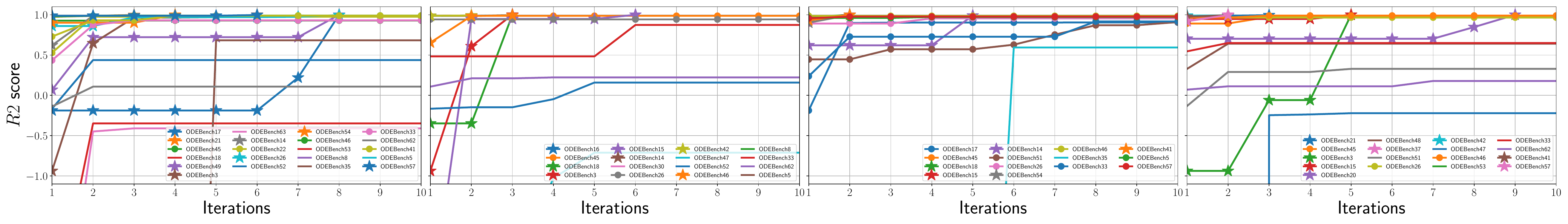}
    \caption{Reflection results over 10 iterations for DYSTS benchmark (top) and ODEBench (bottom).}
    \label{fig:reflection-iter}
\end{figure}

These plots in Figure~\ref{fig:reflection-iter} show the evolution of the $R2$ score over each iteration.
Stars indicate the model reached $R2\geq 0.99$, circles indicate the model reached $0.9\leq R2 < 0.99$.

\end{document}